\documentclass[runningheads,envcountsame,amsfonts]{article}
\usepackage[left=3.75cm]{geometry}
\usepackage{lineno,hyperref}
\usepackage{pdfpages}
\usepackage{amssymb}
\usepackage{float}
\usepackage{pifont}
\usepackage[numbers]{natbib}
\usepackage{tikz}
\usepackage{graphicx}
\usepackage{epstopdf}
\usepackage{url}
\usepackage{amsmath}
\usepackage{amsthm}
\usepackage[utf8]{inputenc}
\usepackage[T1]{fontenc}
\usepackage{lmodern}
\usepackage{ae}
\usepackage{enumerate}
\usepackage{enumitem}
\usepackage{natbib,hyperref}
\newtheorem{theorem}{\bf Theorem}[section]
 \newtheorem{proposit}[theorem]{\bf Proposition}
 \newtheorem{coro}[theorem]{\bf Corollary}
\newtheorem{lem}[theorem]{\bf Lemma}

\newtheorem{notat}{\bf Notation}
\newtheorem{remark}[theorem]{\bf Remark}
 \newtheorem{example}[theorem]{\bf Example}
\newtheorem{condit}{\bf Condition}
\newtheorem{fait}{\bf Claim}

\def\thm#1\par{\medskip\par\noindent\begin{theorem} \strut \sl #1 \end{theorem}\par}
\def\propo#1\par{\medskip\par\noindent\begin{proposit} \strut \sl #1 \end{proposit}
\par}
\def\cor#1\par{\medskip\par\noindent\begin{coro} \strut \sl #1 \end{coro}\par}
\def\lm#1\par{\medskip\par\noindent\begin{lem} \strut \sl #1 \end{lem}\par}
\def\defil#1\par{\medskip\par\noindent\begin{condit} \strut \sl #1 \end{condit}\par}
\def\fct#1\par{\medskip\par\noindent\begin{fait} \strut \sl #1 \end{fait} \cqfd\par}
\def\defi#1\par{\medskip\par\noindent{\begin{defin} \strut  \sl #1 \end{defin}}\par}
\def\nota#1\par{\par\noindent\begin{notat} \nopagebreak  \strut #1  \end{notat}}
\def\rem#1\par{\par\noindent\begin{remark} \nopagebreak \strut \rm #1   \end{remark}}
\def\ex#1\par{\par\noindent\begin{exemple} \nopagebreak \strut #1  \end{exemple}}
\def\cqfd{~~~~~~~~~~$\Box$}

\theoremstyle{plain}

\date{}
    \usepackage[english]{babel}

    \addto{\captionsenglish}{%
       
    }

\title{\bf Variable-Length Codes Independent or Closed with respect to   Edit  Relations}
\author{Jean N\'eraud}

\begin{document}
 
\maketitle
\thanks{{\flushleft Universit\'e de Rouen, Laboratoire d'Informatique, de Traitement de l'Information et des Syst\`emes (LITIS)}, 
 Avenue de l'Universit\'e , 76800 Saint-\'Etienne-du-Rouvray, France.\\
\url{jean.neraud@univ-rouen.fr}
\url{neraud.jean@gmail.com}
\url{neraud.jean.free.fr}\\
\ \\}
\ \\
\begin{abstract}
 We investigate  inference of variable-length codes in  other domains of computer science, such as noisy information transmission or information retrieval-storage:
in such topics,  traditionally  mostly constant-length codewords act. 
The study is relied upon  the two concepts of  independent  and  closed sets:
given  an alphabet $A$ and a binary relation $\tau\subseteq A^*\times A^*$, a set $X\subseteq A^*$
is $\tau$-{\it independent}
 if $ \tau(X)\cap X=\emptyset$; $X$  is  $\tau$-{\it closed} if $\tau(X)\subseteq X$. 
We focus to those word relations whose images  
are computed by applying
 some peculiar combinations of  deletion, insertion, or substitution.
In particular, characterizations of  variable-length codes that are maximal 
 in the families of $\tau$-independent or $\tau$-closed codes are provided.\\
\ \\
\ \\
%
\end{abstract}
%

\section{Introduction}

\label{Introduction}
 In computer science the concept of code is one of the most widely used:
with the terminology of the {\it free monoid}, given some alphabet $A$, a subset $X$ of  $A^*$ (the free monoid generated by $A$) 
is a {\it variable-length code} (for short in the present paper, a {\it code}) if every equation among the {\it words} (or {\it strings}) of $X$ is necessarily trivial.
Famous topics are concerned by such mathematical concept: we particularly mention  the  frameworks of text compression, information transmission, and information storage-retrieval. 

For its part, text compression particularly involves two fundamental concepts from the theory of variable-length codes, namely maximality and completeness \citep[Sec. 3.9]{BPR10}, \citep{H52,ZW77}. 
Given a family of codes over a fixed alphabet $A$, say ${\cal F}$,  a code $X\in{\cal F}$ is {\it maximal} in  ${\cal F}$, if no code in the family can strictly contain $X$.
A set (resp., a code) $X$ is {\it complete} if any word of $A^*$ is a factor of some words of $X^*$, the  submonoid (resp., free submonoid) generated by   $X$:  
actually, a famous result due to Sch\"utzenberger states that, in the family of {\it regular} codes maximality and completeness are two equivalent notions.
In addition,  information transmission by noiseless channels is mostly concerned by variable-length codes.

At the contrary, variable-length codes so far have little impact on the questions related to information transmission by noisy channels or information storage-retrieval.
More precisely, due to technical  specificity, in each of these last topics  only  sets whose elements have a common length,
the so-called {\it uniform} codes, are practically used: this is noticeably illustrated by each of the famous domains of  error-detecting (resp., error-correcting) codes and  Gray sequences.
Numerous outstanding studies have been drawn in such topics:
whereas in the framework of error detection (see e.g. \citep{JK97,KOH02,M05,PW72,S48}) linear algebra appears as a tool of choice,
in the field of Gray codes many questions of interest involve  combinatorics, graph theory and group theory (see e.g.  \citep{CCC92, JWW80, K05,L92,S97}).

However, as is further shown below, in all the preceding domains the part of codes is highlighted thanks to specific notions related to the theory of dependent systems  \citep{JY91}, namely the so-called independent codes, and
 the closed ones. The aim of the present paper, whose a preliminary version appeared in \citep{N20} is to draw some comparative study of the behaviors of such families of codes: this will be particularly done in connection with the two notions of maximality and completeness, which have been introduced above.

\bigskip
-- In the first part of the paper, we  investigate  how  variable-length codes themselves can impact in the framework of noisy information transmission. 
Informally and in very simple terms, with  the notation of the free monoid some model for information transmission  requires two fixed alphabets, say  $A$, $B$: actually 
every information is  modeled by a unique word  $u\in B^*$.  Beforehand, in order to  facilitate the further transmission of that information, usually the word $u$ is transformed in another word $w$ of $A^*$. This is done by making use of a one-to-one {\it coding}   mapping $\phi: B^*\longrightarrow A^*$: 
in numerous cases, $\phi$ consists in an  injective  monoid homomorphism,
whence $X=\phi(B)$ is a variable-length code of $A^*$: such a translation  is particularly illustrated by the well-known examples of the Morse code, or the Huffman code. 
Next, the resulting word $w$ is transmitted via a fixed {\it channel} into some word $w'\in A^*$.
Should $w'$ be altered by some
{\it noise} that is,  $w'$ different from $w$, and then the word $\phi^{-1}(w')\in B^*$ could be different from the initial word $u$.
Therefore,  in order to retrieve $u$, the morphism $\phi$ (thus the code $X$) has to satisfy error-detecting and error-correcting constraints, which of course depend of the channel.
In the most general model of  message transmission, this  channel is represented by some {\it probabilistic transducer}.
However, in the framework of error detection, most of the models only require that highly likely errors need to be taken into account:
in this paper we will  overcome probabilistic aspect that is, we assume the transmission channel modeled by some binary word relation, say $\tau\subseteq A^*\times A^*$. 
To be more precise, every communication process actually  involves the two following main challenges:

\smallbreak
(i)  On a first hand,   in view of minimizing the amount of errors,
some minimum-distance constraint over  $\tau(X)\cup X$  should be applied (with $\tau(X)=\{x': \exists x\in X, (x,x')\in\tau\}$),
the most famous  ones certainly corresponding to the Hamming or the Levenshtein metrics \citep{H50,L65}:
the smaller the distance between the input word $x\in X$ and any corresponding output word $x'\in X\cup\tau(X)$, the more optimal is error detection.\smallbreak
(ii) On another hand, even in case of a noisy transmission, coding the elements of $B^*$, and above all decoding those of $A^*$, 
must allow to retrieve with optimal conditions (especially in terms of time and space) the initial information $u\in B^*$.
From this point of view, according to the nature  itself of information,
numerous  performing families of variable-length codes have been introduced \citep{BPR10,J08}, the most famous one certainly being the family of {\it prefix codes}.
With regard to these families, a fundamental question consists in providing some description of their members, especially from the point of view of maximality and/or completeness.
\citep{BP99,BWZ90,KKK14,L01,L03,N08,ZS95}.

\smallbreak
In the spirit of \citep{JK97,KOH02}, we rely on  {\it dependence systems}:
actually this concept can be associated  with each of the  families of variable-length codes we have just  listed.
Formally, given a set $S$,  a dependence  system consists in  a family ${\cal F}$ of subsets of $S$ satisfying the following property: 
$X$ belongs to ${\cal F}$ if, and only if, some non-empty finite subset  of $X$ exists in ${\cal F}$. Sets in ${\cal F}$ are ${\cal F}$-{\it dependent}, the other ones being ${\cal F}$-{\it independent}.
A famous special  case corresponds to word  binary  relations
$\tau\subseteq A^*\times A^*$, where independent sets are those satisfying 
$\tau(X)\cap X=\emptyset$: we say that they are $\tau$-{\it independent}; similarly sets satisfying $\tau(X)\cap X\neq\emptyset$ are $\tau$-{\it dependent}.
From this point of view, prefix codes are those that are independent with respect to the antireflexive restriction of the famous {\it prefix order}.
Codes that are {\it bifix}, 
or {\it solid}  \citep{BPR10,L01} can similarly be  characterized. 

%
A noticeable fact is that  error-detecting codes are themselves  concerned by dependence systems. 
For that purpose,
consider the family of the relations $\tau$ that can be generated from the so-called {\it basic  edit relations}, which we define below
(given a word $w$, we denote by Subw($w$)  the set of its  subsequences and $|w|$ stands for its length): 

\smallskip
- $\delta_k$, the {\it $k$-character deletion},  associates with every word $w\in A^*$, all the words $w'\in {\rm Subw}(w)$  whose length is $|w|-k$. The {\it at most $p$-character deletion} is $\Delta_p=\bigcup_{1\le k\le p}\delta_k$;

- $\iota_k$, the {\it $k$-character insertion},  is the converse (or inverse) relation of $\delta_k$, moreover we set $I_p=\bigcup_{1\le k\le p} \iota_k$ ({\it at most $p$-character insertion});

-  $\sigma_k$, the {\it $k$-character substitution},  associates with every $w\in A^*$, all $w'\in A^*$ with length $|w|$  such that $w'_i$ (the letter of position $i$ in $w'$), differs from $w_i$ in exactly $k$ positions $i\in [1,|w|]$; we set $\Sigma_p=\bigcup_{1\le k\le p}\sigma_k$.

 \smallskip By applying some combination, one can define other relations: we mention   $S_p=\bigcup_{1\le k\le p}(\delta_1\cup \iota_1)^k$, or $\Lambda_p=\bigcup_{1\le k\le p}(\delta_1\cup \iota_1\cup \sigma_1)^k$.
 For reasons of consistency,  in the whole paper we assume $|A|\ge 2$ and $k\ge 1$.
In addition, in each case we denote by $\underline\tau$ the antireflexive restriction of $\tau$, that is  $\tau\setminus\{(w,w)|w\in A^*\}$. 
Similarly, we denote by $\hat\tau$ the reflexive closure of $\tau$, that is $\tau\cup\{(w,w)|w\in A^*\}$.
\smallskip  For short, we will refer to all these relations as {\it edit relations}.

\smallbreak
Actually,  for every $k\ge 1$, each edit relation $\tau_k\in\{\delta_k,\iota_k,\sigma_k,\Delta_k,I_k,\Sigma_k,S_k,\Lambda_k\}$  leads to introduce a corresponding topology.
For this purpose,
consider the  mapping $d:A^*\times A^*\longrightarrow {\mathbb R}_+$ defined by $d(u,v)=0$ if $u=v$, and  $d(u,v)=\min\{k|(u,v)\in\tau_k\}$ otherwise.
 Although $d$ can be only a partial mapping, in the case where symmetry is ensured
(that is, $\tau_k\in\{\sigma_k,\Sigma_k,S_k,\Lambda_k\}$), it is commonly referred to as {\it metric}, and  otherwise to as  {\it quasi metric} -- for short, in any case we write {\it (quasi) metric}. 
With the preceding definition, the set $X$ is $\tau$-independent if, and only if, for each pair of different words $x,y\in X$,   in the case where the integer $d(x,y)$ is defined, it is necessarily greater than $k$: 
in other words,  with respect to the channel $\tau$, the code $X$ is capable to detect at most $k$-errors.
A natural question consists in investigating the mathematical structure of those  independent codes, in particular as regards  maximality. In our paper, we establish the following result:
{\flushleft {\bf Theorem A.}}
{\it With the preceding notation, let 
$A$ be a finite alphabet, $k\ge 1$ and let  $\tau$ in $\{\delta_k,\iota_k,\sigma_k, \Delta_k, I_k,\Sigma_k,\underline{S_k}, {\underline \Lambda}_k\}$.
Given a regular $\tau$-independent code $X\subseteq A^*$, $X$ is maximal in the family of $\tau$-independent codes if, and only if, it is  complete. 
}

\medskip
In other words, with respect to maximality, codes that are capable to detect at most $k$ errors behave similarly in several of those families of variable-length codes we mentioned above.
This leads us to formulate,  in terms of  word binary relations and variable-length codes, some  specification as regards  error detection (correction).
In addition, in the case where $X$ is assumed to be regular, some corresponding decidability results are stated.

\bigskip 
-- In the second part of our paper we focus to the so-called notion of set closed under a given  word binary relation; in fact it consists in some special condition related to dependence.
Actually, in the literature  several different notions of closed sets can be encountered, the best-known being related to topology or universal algebra \citep{C1965}. 
The concept we refer in the paper is  different:
given a binary relation $\tau\subseteq A^*\times A^*$, a set $X\subseteq A^*$ is {\it closed} under $\tau$ ($\tau$-closed for short) if we have $\tau(X)\subseteq X$.

Beforehand, we  notice a property that will be of a common use in the paper: any non-empty  set is $\tau$-closed if, and only if, it is closed under $\tau^*=\bigcup_{i\in {\mathbb N}} \tau^{i}$.
As such,  many famous  topics are concerned: 
in the  case where the binary relation is some (anti)-automorphism, the so-called {\it invariant} sets \citep{NS20} are directly involved. 
The topics of $L$-systems \citep{RS80}, or congruences in the free monoid \citep{Ni71},
as well as applications to  DNA computing \citep{KPTY97}, are also concerned.
By definition, closed codes cannot have a real impact on error correction, which itself involves independence. 
With the preceding notation, given  some edit relation $\tau_k\in\{\delta_k,\iota_k,\sigma_k,\Delta_k,I_k,\Sigma_k,S_k,\Lambda_k\}$ and its corresponding (quasi) metric $d$, 
 a set $X$ is $\tau_k$-closed if, for every pair of words $x\in X$, $y\in A^*$, the condition $d(x,y)\le k$ implies $y\in X$.
In other words, with respect to $d$, the set $X$ necessarily contains every neighboring word from each of its elements; 
in addition, from the fact that $X$ is also $\tau_k^*$-closed, all its elements  can be  generated in this way. From this last point of view,  the so-called Gray sequences, 
which are  closely connected to information storage-retrieval, are involved.

Given some edit relation, our aim is to characterize the family of corresponding closed  codes. 
In our paper we prove  that, for any $k\ge 1$ there are only finitely many $\delta_k$-closed codes, each of them being itself finite. 
Furthermore, we can  decide whether a given non-complete $\delta_k$-closed code can be embedded into some complete one. 
 We also  prove  that  no closed code can exist with respect to the relations $\iota_k$, nor $\Delta_k$, $I_k$, $S_k$, $\Lambda_k$.

 With regard  to substitutions, given a word $w$,  beforehand we focus to the structure of the set $\sigma_k^*(w)=\bigcup_{i\in {\mathbb N}}\sigma_k^i$.
Actually, excepted for two special cases (that is, $k=1$ \citep{E73,S97}, or $k=2$ with $|A|=2$ \citep[ex. 8, p.77]{K05}), to our best knowledge, in the literature no general description appears.
In any event we provide such a description;
furthermore we establish the following result:
{\flushleft {\bf Theorem B}.}
{\it Let $A$ be a finite alphabet and $k\ge 1$. Given  a complete $\sigma_k$-closed code $X\subseteq A^*$,
either  every word in $X$ has length not greater than $k$, or a unique integer $n\geq k+1$ exists such that $X=A^n$.
In addition for every $\Sigma_k$-closed code $X$, some positive integer $n$ exists such that $X=A^n$.}

\medskip
In other words, no $\sigma_k$-closed code can simultaneously possess words in $A^{\le k}=\bigcup_{0\le i\le k} A^i$
and words in $A^{\ge k+1}=\bigcup_{i\ge k+1}A^i$. 
As a consequence, one can decide whether a given non-complete $\sigma_k$-closed code $X\subseteq A^*$  can be  embedded into some complete one.\\

We now shortly describe the contents of the paper:

-- Section \ref{Sec2} is devoted to the preliminaries. The terminology of the free monoid is settled, moreover we recall two main results from the variable-length code theory: they shall be applied in the sequel.
In addition, in order to further examine the decidability of some questions,
 we review some of the main properties of the so-called regular, and recognizable subsets of $A^*\times A^*$. 

-- In  Section \ref{Sec3} we draw some investigation of variable-length codes that are independent with respect to  some edit relation.
Although it is known that edit relations are regular, we prove that no edit relation can be recognizable.
We also establish Theorem A: the proof lays upon the construction of some word with peculiar properties as regarding edit relations. 

-- Section \ref{Sec4} is devoted to some discussion  over the involvement  of independent  variable-length codes as regards  error detection or error correction. Such a perspective  is  illustrated by  significant examples.
Some decidability results are also stated: they concern the class of regular codes.

-- Codes that are closed under deletion or insertion are studied in Section \ref{deletion-insertion}. 

-- In Section \ref{Sec5}, after having described the structure of $\sigma_k$-closed codes, we prove Theorem~B.
Some algorithmic interpretation is also drawn. 

-- At least, Section \ref{Last} is devoted to some future lines of research related to the present study.
\section{Preliminaries}
\label{Sec2}
Several definitions and notations from the free monoid theory have been fixed above.
The {\it empty word}, denoted by  $\varepsilon$ stands for the word with length $0$.
Given a word $w$,  we denote by $|w|_a$  the number of occurrences of the letter $a$ in $w$.
 Given $t\in A^*$ and $w\in A^+$, we say that $t$ is a {\it factor} ({\it prefix}, {\it suffix})  of $w$ if words $u,v$ exist such that $w=utv$ ($=tv$, $=ut$). 
A pair of words $w,w'$ is {\it overlapping-free} if no pair  $u,v$ exist such that either $uw=w'v$ with $1\le |u| \le |w'|-1$,
or $uw'=wv$
with $1\le |u| \le |w|-1$. 
With such a condition, if $w=w'$, we say that $w$ itself is overlapping-free. Given a subset $X$ of $A^*$, we denote by ${\rm F}(X)$ the set of the factors of $X$ that is, $\{w\in A^*|A^*wA^*\cap X\neq\emptyset\}$.
\subsection{Variable-length codes}
 It is assumed that the reader has a fundamental understanding  with the main concepts of the theory of variable-length codes: we  suggest, if necessary,  that  he (she) refers to \citep{BPR10}.

Given a subset  $X$ of $A^*$, and $w\in X^*$, let $x_1,\cdots,x_n\in X$ such that  $w$
is  the result of the concatenation of  the words $x_1$, $x_{2}$, \dots, $x_n$, in this order. In view of specifying the factorization of $w$ over $X$, we use the notation $w=(x_1)(x_2)\cdots(x_n)$, or equivalently: $w=x_1\cdot x_2\cdots x_n$. For instance, over the set $X=\{a,ab,ba\}$, the word $aba\in X^*$ can be factorized as $(ab)(a)$ or $(a)(ba)$ (equivalently denoted by $ab\cdot a$ or $a\cdot ba$).

A set $X$ is a {\it variable-length code} (a {\it code} for short) if  for any pair of finite sequences of words in $X$, say  $(x_i)_{1\le i\le n}$, $(y_j)_{1\le j\le p}$, the equation
$x_1\cdots x_n=y_1\cdots y_p$ implies  $n=p$, and $x_i=y_i$ for each integer $i\in [1,n]$ (equivalently the submonoid $X^*$ is {\it free}).
In other words, every element of $X^*$ has a unique factorization over $X$.
Given a finite or regular set $X$,  the famous Sardinas and Patterson   algorithm allows  to decide whether or not $X$ is a code.
Since it will be applied several times through the examples of the paper,
it is convenient to shortly recall it. Actually, some ultimately periodic sequence of sets, namely $(U_n)_{n\ge 0}$, is computed, as indicated in the following:
\begin{eqnarray}
\label{SP}
U_0=X^{-1}X\setminus\{\varepsilon\}~~~{\rm and:}~~~
(\forall n\ge 0)~~~U_{n+1}=U_n^{-1}X\cup X^{-1}U_n.
\end{eqnarray}
The algorithm necessarily stops. This corresponds to either $\varepsilon\in U_n$ or  $U_n=U_p$, for some pair of different integers $p<n$: 
 $X$ is  a code if, and only if,  the second condition holds.
A code $X\subseteq A^*$ is {\it prefix} if $X\cap XA^+=\emptyset$ that is,  $U_0=\emptyset$. In addition, $X$ is {\it  suffix} if $X\cap A^+X=\emptyset$ and  $X$ is {\it bifix} if it is both prefix and suffix. 

A positive {\it Bernoulli distribution} consists in some total mapping $\mu$ from $A$ into the set  ${\mathbb R}_+$ of the non-negative real numbers,  such that the equation  $\sum_{a\in A}\mu(a)=1$ holds.
It can be extended into a unique  morphism of monoids from $A^*$ into $({\mathbb R}_+,\times)$,
 which is itself extended into a unique  positive measure  $\mu:2^{A^*}\longrightarrow {\mathbb R}_+$, as indicated is the following:
for each word $w\in A^*$, we set  $\mu(\{w\})=\mu(w)$; in addition, given two disjoint subsets $X,Y$ of $A^*$, we set 
$\mu(X\cup Y)=\mu(X)+\mu(Y)$.
Over a finite alphabet $A$, the corresponding {\it uniform} Bernoulli measure  is defined by $\mu(a)=1/|A|$, for each $a\in A$.
\begin{theorem}{\rm Sch\"utzenberger \citep[Theorem 2.5.16]{BPR10}}
\label{classic}
Let $X\subseteq A^*$ be a regular code.
Then the following properties are equivalent:

{\rm (i)} $X$ is complete;

{\rm (ii)} $X$ is a maximal code;

{\rm (iii)} a positive Bernoulli distribution $\mu$ exists such that $\mu(X)=1$;

{\rm (iv)} for every positive Bernoulli distribution $\mu$ we have $\mu(X)=1$.
\end{theorem}
Actually, this result have been extended to several families of codes, the most famous of which being  those of prefix or bifix codes. 

Another challenging question focuses on methods for embedding a given code $X$ into some maximal one in a given family.
From this point of view,  the following statement answers a question that was beforehand formulated in \citep{R77}:
\begin{theorem}{\rm  \citep{ER85}}
\label{EhrRoz}
Given a non-complete code $X$, let $w\in A^*\setminus {\rm F}(X^*)$ be an overlapping-free word and  $U=A^*\setminus (X^*\cup A^*wA^*)$. Then $Y=X\cup w(Uw)^*$ is a complete code.
\end{theorem}
\subsection{Regular relations, recognizable relations}
\label{Reg-rec}
We assume  the reader to be familiar  with the  theory of regular relations:  if necessary,  we suggest that  he (she) refers to \citep[Chap. II, IV]{S03}.

-- Given a pair of relations $\tau,\rho\in A^*\times A^*$, we denote by $\tau\cdot\rho$ the composition of $\tau$ by $\rho$ that is,
for any $w\in A^*$ we have $\tau\cdot\rho(w)=\rho\left(\tau(w)\right)$; moreover we denote by
 $\overline\tau$ the complement of $\tau$, i.e.  $(A^*\times A^*)\setminus \tau$. 

-- Given a monoid $M$, a family  ${\cal F}$ of subsets of  $M$ is {\it regularly closed} (or equivalently, {\it rationally closed}) if for every pair $X,Y\in{\cal F}$, necessarily each of the three sets  $X\cup Y$, $XY$, and $X^*$ belongs to ${\cal F}$.
Given a family of subsets of $M$, say ${\cal F}$,  its  {\it regular closure} is the smallest   (with respect to the sets inclusion) regularly closed family of subsets of $M$ containing ${\cal F}$.
With such definitions, given two monoids $M$, $N$, a relation $\tau\subseteq M\times N$ is {\it regular} (or equivalently, {\it rational})
 if it belongs to the regular closure of the finite subsets of $M\times N$.

-- A binary relation $\tau\subseteq A^*\times A^*$ is regular  if, and only if, it is the behavior of some finite automaton  with transitions in $A^*\times A^*$.
Equivalently, $\tau$ is the behavior of some finite automaton in {\it normal form} that is, whose transitions belong to $(A\cup\{\varepsilon\})\times( A\cup\{\varepsilon\})\setminus\{(\varepsilon,\varepsilon)\}$ (see e.g. \citep{EM65} or \citep[Sect. IV.1.2]{S03}).

 -- The family of regular relations is closed under union, reverse and composition \citep{EM65,S03}: this can be easily translated in terms of finite automata.

-- The so-called recognizable  relations constitute a noticeable subfamily in regular relations:
a  subset $R\subseteq A^*\times A^*$ is {\it  recognizable} if, and only if, 
we have $R=R\cdot \phi\cdot\phi^{-1}$,
 for some morphism of monoids $\phi:A^*\times A^*\longrightarrow M$, where $M$ is a finite monoid. Equivalently, 
$R$ is the behavior of some finite automaton  with set of states $S$, and where the transitions are done by some {\it action} that is,
a total function from $S\times (A^*\times A^*)$ into  $S$.
Below, we recall a noticeable property, which is commonly attributed to Mezei: it states  a performing characterization of recognizability for the set $R$:
\begin{theorem}
\label{M}
{\rm  \citep[Corollary II.2.20]{S03}}
Given two alphabets $A$, $B$,  and $R\subseteq A^*\times B^*$, the set $R$  is recognizable if, and only if, a finite family
 $\{T_i\}_{i\in I}$ of recognizable subsets of $A^*$ and a finite family $\{U_i\}_{i\in I}$ of recognizable subsets of $B^*$
exist such that $R=\bigcup_{i\in I}T_i\times U_i$.
\end{theorem}
Actually, this result was originally  stated in the  framework of the direct product of two arbitrary monoids.

-- Recognizable relations are closed under composition, complement and intersection,  the intersection with a regular relation being itself regular.

--As a corollary of Theorem \ref{M}, if $X$ is a regular (equivalently recognizable) subset of $A^*$, the relation $X\times X$ is recognizable; see also  \citep[Example II.3.2]{S03} for a corresponding normalized automaton.


-- The relation $id_{A^*}=\{(w,w)|w\in A^*\}$ and its complement $\overline {id_{A^*}}$ are regular. However,
According to Theorem \ref{M},  $id_{A^*}$ is not recognizable  and thus neither is $\overline{id_{A^*}}$. For every regular set $X\subseteq A^*$, the relation $id_X\subseteq A^*\times A^*$ is regular: indeed, we have $id_X=(X\times X)\cap id_{A^*}$, thus $id_X$ is the intersection of a recognizable relation with a regular one.

-- The following result is a consequence of a characterization of regular relations due to Nivat:
\begin{proposit}
\label{MM}
{\rm  \citep[Corollary IV.1.3]{S03}}
Given a regular relation $\tau\subseteq A^*\times A^*$, for every regular subset $X\subseteq A^*$ the set $\tau(X)$ is regular.
\end{proposit}
-- As indicated above, union and composition of regular relations can be translated in terms of finite automata. 
Based on this fact,  given an edit relation $\tau\subseteq A^*\times A^*$, a  finite automaton in normal form with behavior is $\tau$ can actually be constructed.
In other words, the following result holds:
\begin{proposit}
\label{N}
{\rm \citep[Proposition 10]{K02}}
Given a finite alphabet  $A$, every edit relation in\\ $\{\delta_k,\iota_k,\sigma_k,\Delta_k,I_k,\Sigma_k,S_k,\Lambda_k\}$ is regular. 
\end{proposit}
To be more precise, the construction we refered  above lays upon some combination of three basic two-state automata, with respective behavior $\delta_1$, $\iota_1$ or  $\sigma_1$.
For instance, as illustrated by Figure 1, a finite automaton with behavior $\delta_2$, can be obtained by starting with the basic automaton with behavior $\delta_1$ and one duplicate;
then  the terminal state of the first automaton is identified with the initial state of the second one.
\begin{figure}
\begin{center}
\label{Figure1}
\includegraphics[width=12cm,height=12.25cm]{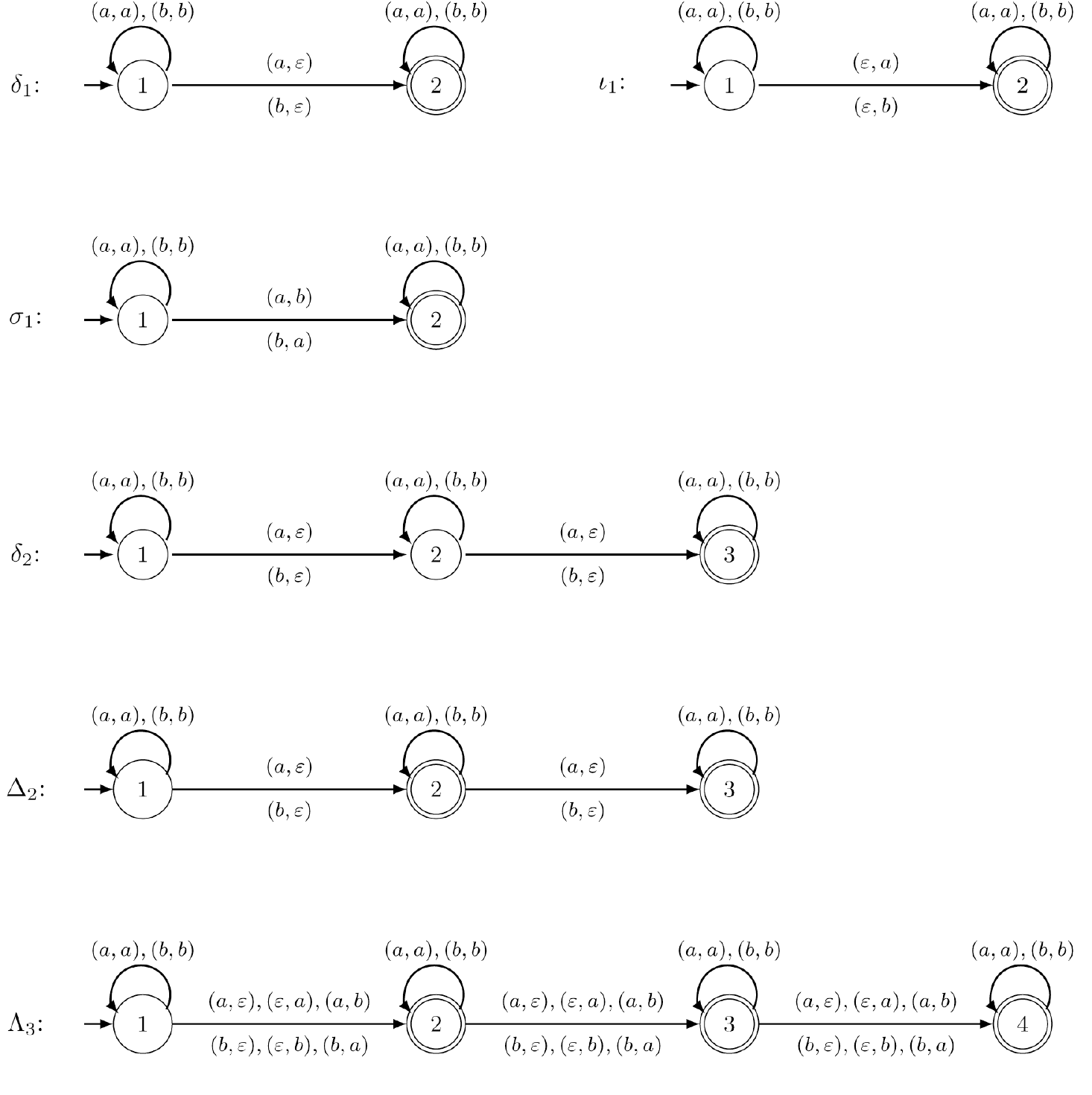}
\end{center}
\caption[]{Over the alphabet $A=\{a,b\}$, automata with behavior $\delta_1$, $\iota_1$, $\sigma_1$, $\delta_2$, $\Delta_2$, $\Lambda_3$.}
\end{figure}
\section{Variable-length codes independent with respect to edit relations}
\label{Sec3}
We start with some general considerations.  
At first, it is straightforward to prove that
 $X$ is  $\tau$-independent if, and only if, it is  independent  with respect to $\tau^{-1}$, the converse relation of $\tau$. 
As regard recognizablity, in view of Proposition \ref{N}, the following result brings some additional property:
\begin{proposit}
\label{N1}
Given a finite alphabet  $A$, every edit relation into $A^*$ is non-recognizable.
\end{proposit}
%
{\flushleft {\it Proof}}
Let $\tau$ be an edit relation into $A^*$. Beforehand we notice that, by definition for very word $w\in A^*$ both the sets $\tau(w)$ and $\tau^{-1}(w)$ are finite.
In addition, some integer $k$ exists such that we have  $\tau(w)\neq\emptyset$ for every word $w\in A^{\ge k}$; therefore $\tau$ itself is necessarily an infinite subset of $A^*\times A^*$.

By contradiction, we assume $\tau$ recognizable.
According to Theorem \ref{M}, two finite families of recognizable subsets of $A^*$, namely  $\{T_i\}_{i\in I}$
and  $\{U_i\}_{i\in I}$ exist such that the equation $\tau=\bigcup_{i\in I}T_i\times U_i$ holds.
Firstly, consider an arbitrary index $i\in I$ and let $w_i\in T_i$. 
It follows from $T_i\times U_i\subseteq\tau$ that we have  $w'\in \tau(w_i)$ for every word $w'\in U_i$. 
This implies $U_i\subseteq\tau(w_i)$, thus $U_i$ being a finite set. As a consequence, since $I$ is finite, the set $U=\bigcup_{i\in I}U_i$ is necessarily finite.
Secondly, from the fact that we have $\tau=\bigcup_{i\in I}T_i\times U_i$, for each $i\in I$ the inclusion $T_i\subseteq\tau^{-1}\left(U\right)$ holds. 
Consequently $T_i$ is a  finite set, hence $\tau$ itself  is actually a finite subset of $A^*\times A^*$: this contradicts the fact that it is an edit relation. Consequently, $\tau$ cannot be recognizable.
\cqfd
\ \\

In \citep[Theorem 10.4]{JK97}, the authors prove that, given a dependence system,  every independent set can be embedded into some maximal one: 
actually, we notice that a similar result holds for  independent codes, that is:

\begin{lem}
\label{embed0}
Given  a binary relation $\tau$ onto $A^*$, every $\tau$-independent code can be embedded into some maximal one.  
\end{lem}
%
{\flushleft {\it Proof}}
%
Let $X\subseteq A^*$ be  a $\tau$-independent code. 
In view of Zorn's lemma, we consider a chain of $\tau$-independent codes containing $X$,  namely ${\cal C}$,
such that ${\cal C}$ is  totally ordered by the sets inclusion:
let   $\hat{X}=\bigcup_{X\in{\cal C}}X$ its least upper bound.
By construction $X$ is included in the set  $\hat{X}$, which is necessarily a code (see e.g. \citep[Proposition 2.1.14]{BPR10}).

By contradiction, assume  that $\hat{X}$ is  $\tau$-dependent  and let  $y\in \hat{X}$ such that $\tau(y)\in\hat{X}$.
By definition, a pair of sets $Y$, $Z$ exist in ${\cal C}$ such that $y\in Y$ and $\tau(y)\in Z$. 
From the fact that ${\cal C}$ is totally ordered by the sets inclusion we have either $Z\subseteq Y$ or $Y\subsetneq Z$.
Actually, since $Y$ is $\tau$-independent, we have $\tau(y)\in Z\setminus Y$, whence necessarily only the inclusion $Y\subsetneq Z$ holds.
But this implies $y,\tau(y)\in Z$: a contradiction with  $Z$ being $\tau$-independent.
Therefore for every word $y\in\hat{X}$, we have $\tau(y)\notin\hat{X}$ that is, $\hat{X}$ is $\tau$-independent.
As a consequence, $\hat{X}$ belongs to ${\cal C}$: this completes the proof.
\cqfd

\ \\
Unfortunately, no more than in \citep{JK97}, no any  method allowing to embed a given $\tau$-independent code into  some maximal one, as for instance provided  by Theorem \ref{EhrRoz}, is actually profiled by Lemma \ref{embed0}.
In the present section, our aim is to establish some characterization of codes  that are maximal in the family of those that are independent with respect to some fixed  edit relation.
We start by constructing a peculiar word:
\begin{lem}
\label{incompletable-del-ins}
Let $k\ge1$, $i\in [1,k]$, $\tau\in\{\delta_i,\iota_i,\sigma_i\}$.
Given a  non-complete
 code $X\subseteq A^*$ an overlapping-free word $w\in A^*\setminus {\rm F}(X^*)$ exists such that the two following conditions hold:

{\rm (i)} $\tau(w)\cap X=\emptyset$;

{\rm (ii)} $w\notin\tau(X)$.
\end{lem}
%
{\flushleft {\it Proof}}
Let $X$ be a non-complete code, and let $v\in A^*\setminus {\rm {\rm F}(}X^*)$. Trivially, we have $v^{k+1}\notin {\rm {\rm F}(}X^*)$.
Moreover, in a classical way a word $u\in A^*$ exists such that $w=v^{k+1}u$ is overlapping-free (see e.g. \citep[Proposition 1.3.6]{BPR10}).
Since we assume $i\in [1,k]$, each word in $\tau(y)$ is constructed by  deleting (inserting, substituting)  at most  $k$ letters from  $w$, hence by construction it contains at least one occurrence of $v$ as a factor.
This implies  $\tau(w)\cap{\rm {\rm F}(}X^*)=\emptyset$,
thus  $\tau(w)\cap X=\emptyset$.

By contradiction, assume that a word $x\in X$ exists such that $w\in\tau(x)$.
It follows from $\delta_k^{-1}=\iota_k$ and $\sigma_k^{-1}=\sigma_k$ that  $w=v^{k+1}u$ is obtained by deleting (inserting, substituting)  at most
 $k$ letters from $x$. Therefore at least one occurrence of $v$ appears as a factor of $x\in F(X^*)$: a  contradiction with $v\notin {\rm {\rm F}(}X^*)$. This implies  $w\notin\tau(X)$.
\cqfd

\ \\
%
%
%
%
As a consequence, we obtain the following result:
\begin{theorem}
\label{classic4}
Let  $k\ge 1$ and  $\tau\in\{\delta_k,\iota_k,\sigma_k\}$.
Given a regular $\tau$-independent code $X\subseteq A^*$, the following conditions are equivalent:

{\rm (i)} $X$ is a maximal code;

{\rm (ii)} $X$ is maximal in the family of $\tau$-independent codes;

{\rm (iii)} $X$ is complete.
\end{theorem}
%
{\flushleft {\it Proof}}
According to Theorem \ref{classic}, every complete $\tau$-independent code is a maximal code, hence it is maximal in the family of $\tau$-independent codes. Consequently, Condition (iii) implies Condition (i), which itself implies Condition (ii).

For proving that Condition (ii) implies Condition (iii), we make use of the contrapositive.
Let $X$ be a non-complete $\tau$-independent code, and let $w\in A^*\setminus {\rm {\rm F}(}X^*)$ satisfying the conditions of Lemma \ref{incompletable-del-ins}.
With the notation of Theorem \ref{EhrRoz},
necessarily  $X\cup\{w\}$, which is a subset of  $Y=X\cup w(Uw)^*$, is a code. 
According to Lemma \ref{incompletable-del-ins}, we have  $\tau(w)\cap X=\tau(X)\cap \{w\}=\emptyset$. Since  $X$ is $\tau$-independent and $\tau$ antireflexive,
 this implies $\tau(X\cup\{w\})\cap (X\cup\{w\})=\emptyset$, thus $X$ non-maximal as a $\tau$-independent code.
\cqfd
%

\ \\
We note that, for $k\ge 2$ no $\Lambda_k$-independent  set can exist:
indeed, we have $x\in\sigma_1^2(x)\subseteq \Lambda_k(x)$.
Similarly, it follows from $x\in\delta_1\iota_1(x)\subseteq (\delta_1\cup\iota_1)^2(x)$ that for $k\ge 2$, no $S_k$-independent set can exist: this justifies the introduction of restrictions such as ${\underline \Lambda}_k$ or $\underline S_k$.
On another hand, the following result is a direct consequence of Theorem \ref{classic4}:
\begin{coro}
\label{coro1}
Let  $\tau\in\{\Delta_k,I_k,\Sigma_k, \underline S_k,{\underline \Lambda}_k\}$.
Given a regular $\tau$-independent code $X\subseteq A^*$, the three following conditions are equivalent:

{\rm (i)} $X$ is a maximal code;

{\rm (ii)} $X$ is maximal in the family of $\tau$-independent codes;

{\rm (iii)} $X$ is complete.
\end{coro}
%
{\flushleft {\it Proof}}
As indicated above, if $X$ is complete, it is  a maximal code, thus it is maximal as a $\tau$-independent code.
Consequently, Condition (iii) implies Condition (i), which itself implies Condition (ii).
For proving that Condition (ii) implies Condition (iii),
once more we argue by contrapositive that is, with the notation  of Lemma \ref{incompletable-del-ins},
we prove that $X\cup\{w\}$ remains independent. 
By definition, for each $\tau\in\{\Delta_k,I_k,\Sigma_k,{\underline\Lambda}_k,\underline S_k\}$, we have $\tau\subseteq\bigcup_{1\le i\le k}\tau_i$, with $\tau_i\in\{\delta_i,\iota_i,\sigma_i\}$.
According to Lemma \ref{incompletable-del-ins},  since $\tau_i$ is antireflexive, for each $i\in [1,k]$ we have $(X\cup \{w\})\cap\tau_i(X\cup\{w\})=\emptyset$:
this implies $(X\cup\{w\})\cap\bigcup_{1\le i\le k}\tau_i(X\cup\{w\})=\emptyset$, thus $X\cup\{w\}$ being $\tau$-independent.
\cqfd
%
\section{Independent variable-length codes and  error detection}
\label{Sec4}
As indicated in the Introduction, as regards  information transmission, 
according to the fact that channels are considered noisy or not,
 there have always been historically specific mathematical methodologies for dealing with codes.
In this section,  we intend to investigate how  some aspects of error detection (correction) could be more deeply regarded in the field of the free monoid, and especially the framework of variable-length codes.

\subsection{Error-detection constraints}
\label{EDC}
Let $\tau\subseteq A^*\times A^*$ be some edit relation,  ${\cal F}\subseteq 2^{A^*}$ a family of variable-length codes  and $X\in{\cal F}$.
The goal is to  transmit messages of $X^*$ via the channel $\tau$, by achieving optimal error detection (resp., error correction) in output messages.
For that purpose, several  conditions should be taken into account. 
Among the constraints we state below, the first three ones are retrieved from now classical sources of the  literature (see e.g. \citep{JK97,MWS77}).
All those conditions are consistent with the model of information transmission we fixed above:
this allows some simplicity in their formulation.
There is one point to be made at the outset: according to the context, it could be difficult, if not  impossible,  to satisfy all those conditions:
some compromise should be adopted (nevertheless several constraints appear mandatory).  
Notice that noiseless channels, which involve the classical field of variable-length codes, are actually covered by the whole conditions. 
Recall that, given an edit  relation $\tau$, we denote by  $\underline\tau$ the antireflexive restriction of $\tau$ and by 
 $\hat\tau$ its reflexive closure.
\begin{enumerate}[label={\rm (c\arabic*)}]
\item \label{0} {\it Synchronization constraint}: 

For every input word factorized as $w=(x_1)\cdots (x_n)$ ($x_i\in X$, $1\le i\le n$)  any corresponding output message $w'$ has to be  factorized as $w'\in\hat\tau(x_1)\cdots\hat\tau(x_n)$.
\item \label{2}  $X$ is $\underline\tau$-{\it independent}: $X\cap\underline\tau(X)=\emptyset$.
\item
\label{c}
{\it Error-correction constraint:}

$(\forall x\in X) (\forall y\in X)~~~~\tau(x)\cap\tau(y)\neq\emptyset\Longrightarrow x=y$.

\item \label{1}   {\it $X$ is maximal in the family ${\cal F}$.

\item \label{3}  $\hat\tau(X)$ is a code.

\item \label{4} $\underline\tau(X)$ is a code.
}
\end{enumerate}

\smallskip
In what follows, we discuss  these conditions:
\begin{itemize}[label={\bf $\rightarrow$},leftmargin=* ,parsep=0cm,itemsep=0.2cm,topsep=0.2cm]
\item The so-called synchronization constraint appears mandatory.
Indeed, as illustrated in Example \ref{e1}, it  ensures that, in the case where  the output word $w'$ belongs to $X^*$  
no error occurred. In order to retrieve the factorization of $w'$ over ${\hat\tau}(X)$, as in the example of Morse code, some pause symbol could be inserted  after each factor $x_i\in X$
in  the input word $w=(x_1)\cdots (x_n)$.
\item  
The constraint on independence \ref{2} is crucial:  
as indicated above  it  expresses some characterization of the error-detecting capability of the code $X$, with respect to the channel $\tau$, or equivalently  the corresponding  (quasi) metric  adopted in $A^*$.  
In other words, joined with the synchronization constraint, every $\tau$-independent code $X$ 
is capable to detect at most $k$ errors in any block of $\hat\tau(X)$ from the output message.
\item Condition \ref{c}  states a classical definition of  $\tau$-error correcting codes. 
\item   
 According to Kraft inequality, given a positive Bernoulli measure $\mu$ over $A^*$,
for every  variable-length code $X$ we have $\mu(X)\le 1$. 
According to Theorem \ref{classic}, the condition $\mu(X)=1$ itself is equivalent to $X$ being complete
that is, every word in $A^*$ being actually a factor of some message in $X^*$: for such codes no part of $X^*$ appears spoiled. In addition, the set $X$ is a maximal code, hence it is maximal in ${\cal F}$ \ref{1}: in other words, $X$ cannot be improved with respect to that family (cf.  examples \ref{e0}, \ref{e1}, \ref{e5}).

On another hand, depending on the combinatorial structure of the family ${\cal F}$,  codes that are maximal in ${\cal F}$  need not to be complete:
this is especially  the case for  solid codes or {\it comma-free} codes \citep{L01,L03}, however these codes possess  noticeable importance as regards  decoding.
Given an edit relation $\tau$, the preceding Theorem \ref{classic4} and Corollary \ref{coro1} bring a  characterization of those maximal $\tau$-independent codes which are complete.
\item Condition  \ref{3} arises naturally for $\hat\tau(X)$: it expresses that the factorization of every output message over the set  $\hat\tau(X)=X\cup\underline\tau(X)$ is done in a unique way.
Nevertheless, this constraint  appears very strong. Indeed,  joined with maximality \ref{1} it implies  $\hat\tau(X)=X$: 
since the channel is assumed to satisfy  the synchronization constraint  \ref{0}, actually $\tau$ is  the identity over $A^*$ that is, it represents the noiseless channel.

On another way, lower constraints might be invoked.
From this point of view, we notice that,  even in the case where $\hat\tau(X)$ is not a code,
$X$ can possess some noticeable error correction capability (cf. Example \ref{e2} or Example \ref{e3}).
\item 
Consider some output message $xx'y$, with $x\in X^*$, $x'\in\underline\tau(X)^+$ and $y\in \hat\tau(X)^+$.
 Even if  $\hat\tau(X)$ is not a code, with Condition \ref{4}  the word $x'$ nevertheless has  a unique decomposition over $\underline\tau(X)$.  

Nevertheless, even if that condition is not satisfied, error correction property may fortunately holds,
as attested by Example  \ref{e3}.

\medskip
\end{itemize}
\subsection{A series of examples}
\label{EXEM}
In what follows, in the framework of a binary alphabet $A=\{a,b\}$, we illustrate how various can be the configurations related to some conjunction of the preceding constraints.
\begin{example} {\rm
\label{EEEE1}
\label{e0}
Every maximal uniform code 
is   equal to $A^n$, for some $n\ge 1$. On a first hand, with respect to $\Sigma_k$ and $\underline\Lambda_k$,  such a code is never independent that is, has no error-detecting capability. 
On another hand, for every $k\le n$ the code $A^n$ is independent with respect to $\delta_k$ and $\Delta_k$. Moreover
$A^n$ is independent with respect to $\iota_k$  and $I_k$ for every $k\ge 1$. 
} \end{example}
\begin{example} {\rm
\label{e1} Consider the regular prefix code $X=\{(ba)^n\{a,b^2\}|n\ge 0\}$ (cf. Figure 2). 
In view of
Theorem \ref{classic}, taking for $\mu$ the uniform
Bernoulli distribution over the alphabet $A$ it follows from $\mu(X) = 1$, 
that X is maximal.
For every $n\ge 0$, we have $|X\cap A^n|=1$, hence $X$ is $\sigma_1$-independent. 
With regard to $\delta_1$, we have $\delta_1(\{a,b^2\})=\{\varepsilon,b\}$ and, 
for every $n\ge 1$: $\delta_1\left((ba)^na\right)=\{a(ba)^{n-1}a,b(ba)^{n-1}a,(ba)^n\}$ and 
$\delta_1\left((ba)^nb^2\right)=\{a(ba)^{n-1}b^2,b(ba)^{n-1}b^2,(ba)^nb\} $, therefore
$X$ is $\delta_1$-independent that is,  equivalently it is $\iota_1$-independent: as a consequence, $X$ is $\underline S_1$-independent and $\underline\Lambda_1$-independent.
In addition,  for every $k\ge 1$ and every $x\in X$ we have $|\sigma_k(x)|=|x|$,  
thus $X$ is $\underline\Sigma_k$-independent.

On another hand, taking $w=baa\in X$ as an input message,
via the channel $\underline\Lambda_1$ (resp., $\Sigma_1$) the output message $w'=aaa$ can be returned.
Notice that, with respect to the notation introduced in Section 2.1,  $w'$ itself can be factorized either as $(aaa)\in\underline\Lambda_1(X)$ (resp., $(aaa)\in\Sigma_1(X)$) or $(a)(a)(a)\in X^*$.
With the second factorization, since the Levenshtein  metric between the words $w$ and $a$ is  $2$,
 without the synchronization condition  \ref{0},  no error could be detected  with respect to the channels  $\underline\Lambda_1$. Similarly, since the  Hamming metric between $w$ and $a$ is not defined,   without Condition \ref{0} no error could be detected with respect to $\Sigma_1$.
More precisely, with this condition, with respect to each of the preceding channels, we shall only retain the  factorization $(aaa)$ for $w'$, 
in which  exactly one error may  effectively be detected.
} \end{example}
\begin{figure}
\begin{center}
\label{Figure111}
\includegraphics[width=8cm,height=4cm]{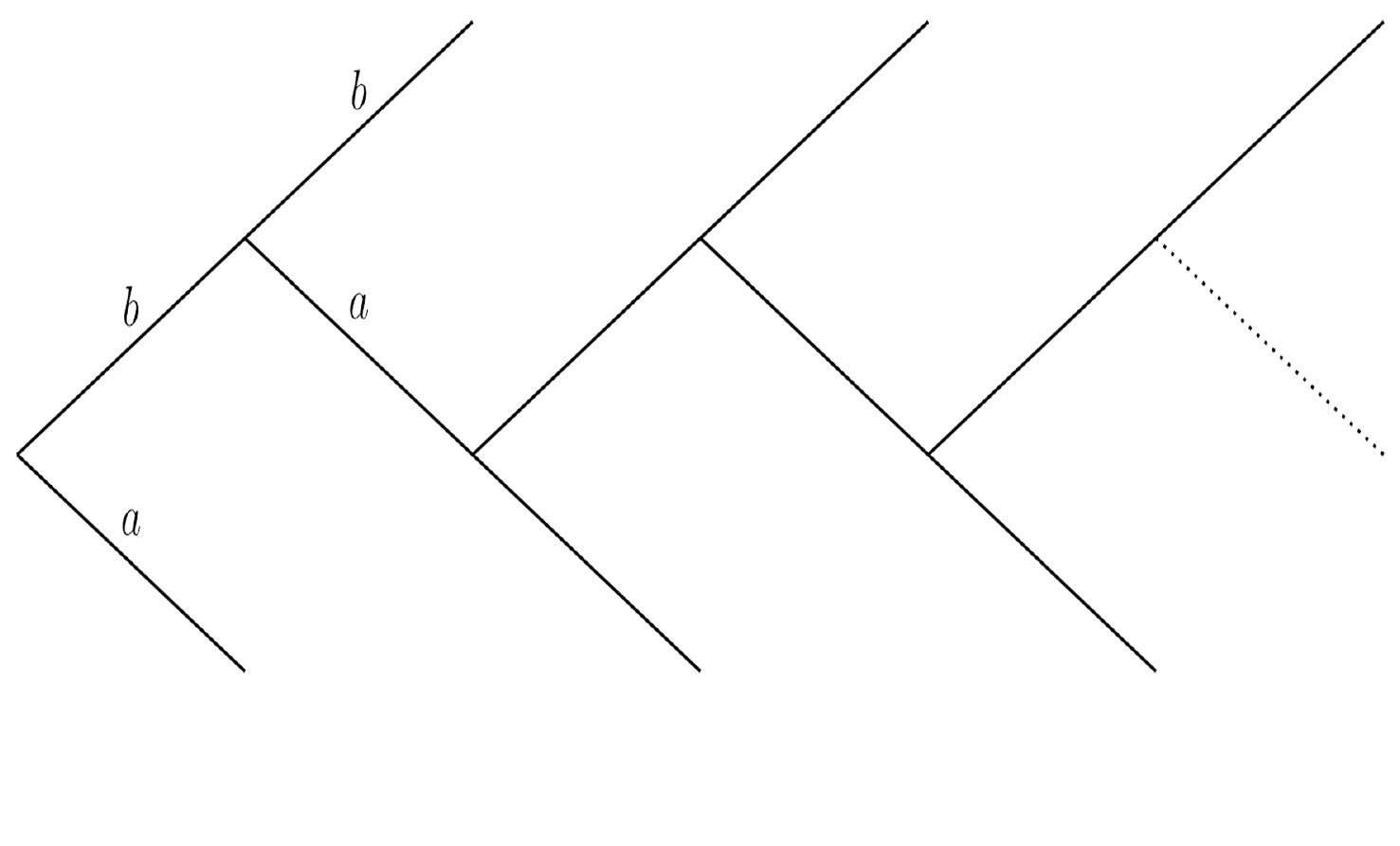}
\end{center}
\caption[]{Example \ref{e1}: A tree-like representation of the infinite maximal prefix code $X=\{(ba)^n\{a,b^2\}|n\ge 0\}$. Elements of $X$ are in one-to-one correspondence with labels of paths from the root to some leaf.}
\end{figure}
\begin{example} {\rm
\label{e5}
 Let $\tau=\sigma_1=\underline\tau$ and $X$ be the bifix code $\bigcup_{n\ge 0}\{ab^na,ba^nb\}$. Taking for $\mu$ the uniform measure we obtain
$\mu(X)=2\cdot1/4\sum_{n\ge 0}(1/2)^n=1$, thus $X$  is  maximal in the family of bifix codes \ref{1}.
Moreover $X$ is $\tau$-independent \ref{2}: indeed $\tau(X)$ is the union of the sets $Y_i$, ($1\le i\le 5$) which as defined as indicated in the following:
\begin{eqnarray}Y_1=\bigcup_{n\ge 1}\{ab^n,b^na\},~~Y_2= \bigcup_{n\ge 1}\{a^nb,ba^n\},~~Y_3=\bigcup_{m,n\ge 1}\{ab^mab^na,ba^mba^nb\},\nonumber
\end{eqnarray}
\begin{eqnarray}
Y_4=\bigcup_{n\ge 0}\{a^2b^na, b^2a^nb\},~~Y_5=\bigcup_{n\ge 0}\{ab^na^2,ba^nb^2\}.\nonumber
\end{eqnarray}
According to Theorem \ref{classic4}, $X$ is maximal in the family of $\sigma_1$-independent codes.
Since we have $ab\in\sigma_1(aa)\cap\sigma_1(bb)$, $X$ does not satisfy the error correction constraint  \ref{c}.
The condition of being a code  is no more satisfied for $\hat\tau(X)$  and $\underline\tau(X)$. Indeed, the following equation holds upon the words of $\tau(X)$:

$(ab^mab^na)( ba^mba^nb)=(ab^m)( ab^n)(ab)( a^mb)(a^nb).$

Actually, given two different words $x,y\in X$, the condition $\tau(x)\cap\tau(y)\neq \emptyset$
implies $\tau(x)\cap\tau(y)=\{ab\}$ or $\tau(x)\cap\tau(y)=\{ba\}$
that is, $\{x,y\}=\{a^2,b^2\}$. As a consequence the bifix code $X\setminus\{a^2,b^2\}$ is error-correcting \ref{c}.
} \end{example}

\begin{example} {\rm  (\citep[Example 4.3]{JK97} extended)
\label{e2}
 Let $\tau=\delta_1=\underline\tau$ and  $X=\{a^m b^n,b^pa^q\}$, with $m,n,p,q\ge 2$. 
We have
$\tau(a^mb^n)=\{a^{m-1}b^n, a^mb^{n-1}\}$ and
$\tau(b^pa^q)=\{b^{p-1}a^q,b^pa^{q-1}\}$, whence $X$ is $\tau$-independent \ref{2}. 

The set $\underline\tau(X)=\tau(X)$ is a (prefix) code \ref{4},
however, as attested by what follows, $\hat\tau(X)$  is not a code.

Consider the input message $w=(a^mb^n)(b^pa^q)(a^mb^n)(b^pa^q)$.
Via the channel $\tau$, the word   $w'=a^mb^{n+p-1}a^{m+q-1}b^{n+p-1}a^q$ may be a returned output message. Actually,
according  to the synchronization constraint \ref{0},  $w'$ may be factorized over $\hat\tau(X)$ in each of the following different ways:

$w'=(a^m b^{n})(b^{p-1}a^q)(a^{m-1}b^n)(b^{p-1}a^q)\in a^m b^n\cdot\tau(b^pa^q) \cdot\tau(a^m b^n)\cdot\tau(b^pa^q),$

$w'=(a^m b^{n-1})(b^{p}a^q)(a^{m-1}b^n)(b^{p-1}a^q)\in \tau(a^m b^n)\cdot b^pa^q\cdot\tau(a^m b^n)\cdot\tau(b^pa^q),$

$w'=(a^m b^{n-1})(b^{p}a^{q-1})(a^{m}b^{n})(b^{p-1}a^q)\in \tau(a^n b^n)\cdot\tau(b^pa^q)\cdot a^m b^n\cdot\tau(b^pa^q),$

$w'=(a^m b^{n-1})(b^{p}a^{q-1})(a^{m}b^{n-1})(b^{p}a^q)\in\tau(a^n b^n)\cdot\tau(b^pa^q)\cdot\tau(a^n b^n)\cdot b^pa^q.$

Since we have $\tau(a^m b^n)\cap\tau(b^pa^q)=\emptyset$, the code $X$ 
is error-correcting with respect to $\tau$  \ref{c}.
Furthermore, in each case, we have:

$w'\in\hat\tau(a^m b^n)\cdot\hat\tau(b^pa^q)\cdot\hat\tau(a^m b^n)\cdot\hat\tau(b^p a^q)$.
} \end{example}

\begin{example} {\rm
\label{e3}
Let  $\tau=\Delta_2$ and  $X$ be the bifix code  $\{a^2b^3,b^4a^2\}$.

We have 
$\underline\tau(a^2b^3)=\{ab^3,a^2b^2,b^3,ab^2,a^2b\}$ and 
$\underline\tau(b^4a^2)=\{b^3a^2,b^4a,b^2a^2, b^3a,b^4\}$, hence $X$ is error-correcting \ref{c}. 
However, $\underline\tau(X)$ is not a code \ref{4}, as attested by the following equation among its elements:

$(a^2b^2)(b^2a^2)(ab^3)(b^2a^2)=(a^2b)(b^3a)(a^2b^2)(b^3a^2).$\\
Nevertheless, we notice that each side of the previous equation belongs to the set:

$\hat\tau(a^2b^3)\cdot\hat\tau(b^4a^2)\cdot\hat\tau(a^2b^3)\cdot\hat\tau(b^4a^2),$\\
hence   the output message $a^2b^4a^3b^5a^2$ may be corrected as  $(a^2b^3)(b^4a^2)(a^2b^3)(b^4a^2)$.
} \end{example}
%
\begin{example} {\rm
\label{e4}
Let $X=\{a^4,a^3b,ab^2,bab\}$. Since we have $\delta_1(X)=\{a^3,a^2b,ab,ba, b^2\}$,
$X$ is $\delta_1$-independent that is,  error-detecting with respect to $\delta_1$ \ref{2}. 
Since we have  $a^2\in\delta_1(a^3)\cap\delta_1(a^2b)$, $X$ is not error-correcting. Notice that $\delta_1(X)$ itself is a (maximal prefix) code \ref{c}.
} \end{example}
%
\begin{example} {\rm
\label{e6}
Let $\tau=\delta_1$
and  $X$ be the non-complete context-free bifix  code $\{a^nb^n|n\ge 2\}$. 
Since we have $\tau(X)=\{a^{n-1}b^n|n\ge 2\}\cup\{a^nb^{n-1}|n\ge 2\}$, the code $X$ is $\tau$-independent \ref{2}.
In addition, since $n\neq m$ implies $\tau(a^nb^n)\cap\tau(a^mb^m)=\emptyset$, $X$ is error-correcting \ref{c}.   

Notice that the set $\underline\tau(X)=\tau(X)$ remains a code \ref{4} which is  bifix and error-detecting with respect to the channel $\tau$.
Indeed, we have $\tau^2(X)=\bigcup_{n\ge 2}\{a^{n-2}b^n,a^{n-1}b^{n-1}, a^{n}b^{n-2}|n\ge 2\}$, thus $\tau\left(\underline\tau(X)\right)\cap\underline\tau(X)=\emptyset$. However $\underline\tau(X)$ is not error-correcting  (we have $a^{n-1}b^{n-1}\in\tau(a^{n-1}b^n)\cap\tau(a^nb^{n-1})$). 

Actually,  $\hat\tau( X)$ is a code \ref{3}. Indeed, by applying Sardinas and Patterson algorithm (cf. \ref{SP})  to $\hat\tau( X)$,
we obtain $U_0=\{b\}$ thus $U_p=\emptyset$ for all $p\ge 1$.
} \end{example}
\subsection{Some decidability results}
\label{decidable0}
As indicated above, the main feature of the synchronization constraint essentially consists in guiding the correction process,
and it could be directly implemented in the channel.
In what follows our aim is to examine whether the condition \ref{2}--\ref{4} can be decidable.
We start by proving a technical property, which  actually holds without assuming that $X$ is a code:
\begin{lem}
\label{decidCondC}
Given a code $X\subseteq A^*$, it  satisfies the error correction constraint if, and only if, for each word $x\in X$, $\tau(x)\neq\emptyset$ implies
$\tau^{-1}\left(\tau(x)\right)\cap X=\{x\}$.
\end{lem}
%
{\flushleft {\it Proof}}
Let $x\in X$ such that $\tau(x)\neq\emptyset$ and let $y\in\tau^{-1}\left(\tau(x)\right)\cap X$.
By construction we have $\tau(x)\cap\tau(y)\neq\emptyset$: if $X$ satisfies the error correction constraint then we obtain  $x=y$, thus $\tau^{-1}(\tau(x))\cap X=\{x\}$.

Conversely, assume that $x\in X$ and $\tau(x)\neq\emptyset$ implies
$\tau^{-1}\left(\tau(x)\right)\cap X=\{x\}$. Let $x,y\in X$ such that $\tau(x)\cap\tau(y)\neq\emptyset$. 
For every  word $y'\in \tau(x)\cap\tau(y)$, necessarily we have $y\in\tau^{-1}(y')\subseteq \tau^{-1}(\tau(x))$: this  implies  $y\in\{x\}$, thus $x=y$, whence $X$ is error-correcting. \cqfd
%

\ \\
The following result provides some decidability properties related to our conditions:
\begin{proposit}
\label{decidable}
Let $A$ be some finite alphabet, and $k\ge 1$. Given a regular variable-length code $X\subseteq A^*$, and given an edit relation $\tau\in\{\delta_k,\iota_k,\sigma_k,\Delta_k,I_k,\underline{S}_k,\underline\Lambda_k\}$, each of the following properties holds:

{\rm (i)} If $X$ is finite then each of the conditions \ref{2}--\ref{4} is decidable.

{\rm (ii)} It can be decided whether $X$ is maximal in the family of $\tau$-independent codes \ref{1} and  whether ${\hat\tau}(X)$ is a code
\ref{3}.

{\rm (iii)} If $\tau$ belongs to $\{\delta_k,\iota_k,\sigma_k,\Delta_k,I_k, S_1,\Lambda_1\}$ ($k\ge 1$) 
then one can decide whether $X$  is $\underline\tau$-independent \ref{2}, and  whether $\underline\tau(X)$ is a code \ref{4}.

\end{proposit}
%
{\flushleft {\it Proof}}
Let $X\subseteq A^*$ be a regular code. We consider one by one  our conditions \ref{2}--\ref{4}:
\begin{itemize}[label=--,leftmargin=*]
\item {\it Condition} \ref{2}
Firstly, assume that $X$ is a finite set. 
Since $\tau$ is an edit relation, $\underline\tau(X)$ is finite, thus $\underline\tau(X)\cap X$ itself is finite: trivially it can be decided whether or not it is the empty set.
Secondly, in the case where $\tau$ belongs to $\{\delta_k,\iota_k,\sigma_k,\Delta_k,I_k, S_1,\Lambda_1\}$ ($k\ge 1$)
we have  $\underline\tau=\tau$,
therefore,  the equation $\underline\tau(X)\cap X=\emptyset$ is equivalent to $\tau\cap(X\times X)=\emptyset$.
As indicated in Section \ref{Reg-rec}, $X\times X$ is a recognizable subset of  $A^*\times A^*$. In addition, according to
Proposition \ref{N} $\tau$ is regular: this implies %
$\tau\cap(X\times X)$ regular, hence it can be decided whether or not it is the empty set, in other words Condition \ref{2} is decidable.
\item {\it Condition} \ref{c} Since $\tau$ is an edit relation,  for any finite subset $X$ of $A^*$,  and for each $x\in X$, the set $\tau^{-1}\left(\tau(x)\right)\cap X$ is necessarily finite.
Therefore, according to Lemma \ref{decidCondC}, one can decide whether $X$ satisfies the error correction condition.
\item
{\it Condition} \ref{1}
 According to Theorem \ref{classic4} and Corollary \ref{coro1},  $X$ is maximal in the family of $\tau$-independent codes if, and only if, it is complete. According to  Theorem \ref{classic} (iii), this is equivalent to $\mu(X)=1$, where $\mu$ stands for  the uniform Bernoulli distribution. Consequently,  maximality in the family of $\tau$-independent codes can be decided for every regular (a fortiori finite) code. 
\item
 {\it Condition} \ref{3} 
By definition, we have $\hat\tau=\tau\cup id_{A^*}$. As indicated in Section \ref{Reg-rec}, the relations $id_{A^*}$ and $\tau$ are regular, therefore their union $\hat\tau$ is regular; 
in addition, since $X$ is regular,  according to Proposition \ref{MM} $\hat\tau (X)$ is regular.  Consequently one can decide whether it is a code by applying Sardinas and Patterson algorithm.
\item
{\it Condition} \ref{4} 
If $X$ is finite,  ${\underline\tau(X)}$ itself is finite: once more it can be decided whether it is a code  by applying Sardinas and Patterson algorithm. 
If  $\tau$ belongs to $\{\delta_k,\iota_k,\sigma_k,\Delta_k,I_k,S_1,\Lambda_1\}$ ($k\ge 1$),
we have $\underline\tau=\tau$. According to Proposition \ref{MM}, since $\underline\tau$ and $X$ are regular, $\underline\tau(X)$  itself is regular: 
once more by applying Sardinas and Patterson algorithm, one can decide whether or not $\underline\tau(X)$ is a code.\cqfd
\end{itemize}
%
In the case where $X$  is not  a finite set, Proposition \ref{decidable}  lets actually open the three following questions:
\begin{enumerate}[label={\rm Q\arabic*)}]
\item 
\label{Q1} Let $k\ge 2$, $\tau\in\{S_k,\Lambda_k\}$. Given a regular  code  $X\subseteq A^*$, is $X$ a $\underline\tau$-independent set,
or equivalently does the equation $(\tau\cap\overline{id_A^*})(X)\cap X=\emptyset$ hold?
%
\item 
\label{Q2} 
Given a regular code $X\subseteq A^*$, does it satisfy  the error correction constraint? Note that according to  Lemma \ref{decidCondC}, this is  equivalent to $(X\times A^*)\cap(\tau\cdot\tau^{-1})\subseteq id_{A^*}$
that is, $(X\times A^*)\cap(\tau\cdot\tau^{-1})\cap\overline{id_{A^*}}=\emptyset$.
\item 
\label{Q3} Let $k\ge 2$, $\tau\in\{S_k,\Lambda_k\}$. Given a regular code $X\subseteq A^*$ is  the set ${\underline\tau(X)}=(\tau\cap\overline{id_{A^*}})(X)$ a variable-length code?
\end{enumerate}
Since $\overline{id_{A^*}}$ is not recognizable and since, in the most general case, intersection of sets is not regularity preserving,  none of the preceding questions is presently known to be decidable.\\

\medskip
As indicated in the Introduction, the second part of the paper is devoted to investigating the behavior of edit relations with regard to closed sets. We will start with the relations $\delta_k$, $\iota_k$, $\Delta_k$, $I_k$, $S_k$.
\section{Codes closed under deletion or insertion}
\label{deletion-insertion}
Recall that, given  a relation $\tau\subseteq A^*\times A^*$, a set $X\subseteq A^*$ is $\tau$-closed if $\tau(X)\subseteq X$.
We start with  some general  properties of closed codes. Firstly, the following result  comes from the definition: actually it will be frequently applied in the sequel.
\begin{lem}
\label{iteration-tau-sur-X}
Let $\tau\in A^*\times A^*$ and $X\subseteq A^*$. 
Then $X$ is $\tau$-closed if, and only if, it is $\tau^*$-closed.
\end{lem}
%
{\flushleft {\it Proof}}
Assume that $X$ is $\tau$-closed. For each $i\in {\mathbb N}$ we have $\tau^{i+1}(X)=\tau(\tau^i(X))$ therefefore, by induction over $i\ge 0$ we obtain $\tau^i(X)\subseteq X$, thus $\tau^*(X)\subseteq X$.
Conversely, by definition   $\tau^*=\bigcup_{i\in {\mathbb N}}\tau^i$ implies  $\tau(X)\subseteq\tau^*(X)$, whence
 $X$  being $\tau^*$-closed implies $\tau(X)\subseteq X$. 
\cqfd
%

\ \\
Secondly, as regards maximality, the following result states that closed codes have a behavior quite similar to that of independent codes.
\begin{lem}
\label{embedding-tau-maximal}
Given a binary relation $\tau$ onto $A^*$, every $\tau$-closed code can be embedded into some maximal one.
\end{lem}
%
{\flushleft {\it Proof}}
In a classical way, we apply Zorn's lemma. 
Let ${\cal C}$ be a chain ordered by inclusion of $\tau$-closed codes  and let  $\hat{X}=\bigcup_{X\in{\cal C}}X$. 
By construction  the set  $\hat{X}$ is necessarily a code \citep[Proposition 2.1.14]{BPR10}.
For proving that it is $\tau$-closed, we consider a word $x\in \hat{X}$ that is, $x\in X$ for some $X\in {\cal C}$.
Since $X$ is $\tau$-closed, we have $\tau(x)\subseteq X$, thus $\tau(x)\subseteq {\hat X}$.
\cqfd
%

\ \\
As in the case of independence,  the preceding property only states a condition of existence. 
In other words,  it  unfortunately does not allow to implement any practical method for embedding a non-maximal code into some maximal one: actually the question of developing such method remains open. However, in the special case of $\delta_k$-closed codes, we will see that such a procedure can be obtained (cf. Corollary \ref{coro-delta}).
\begin{remark} {\rm
In the literature,   in the framework of dependence systems \citep{C1965} another notion of closed set appears:
for instance, with regard  to the prefix order $P$,  such sets correspond to unitary submonoids of $A^*$. 
The two notions do not intersect: indeed in the sense of our paper, unitary submonoids  are not $P$-closed. }
\end{remark}

Next we focus to  $\delta_k$-closed codes.
A noticeable fact is that corresponding closed codes are necessarily finite, as attested by the following result:
\begin{proposit}
\label{complete-delta-closed}
Given  a $\delta_k$-closed code $X$, and 
$x\in X$, we have $|x|\in [1,k^2-k-1]\setminus\{k\}$.
\end{proposit}
%
{\flushleft {\it Proof}}
It follows from $\varepsilon\notin X$ and $X$ being $\delta_k$-closed that $|x|\neq k$.
By contradiction, assume $|x|\ge (k-1)k$ and
let $q,r$ be the unique pair of integers such that $|x|=qk+r$, with $0\le r\le k-1$. 
Since we have $0\le rk\le (k-1)k\le |x|$,  an integer $s\geq 0$ exists such that $|x|=rk+s$,
thus words $x_1,\cdots, x_k,y$ exist such that 
$x=x_1\cdots x_ky$, with  
$|x_1|=\cdots=|x_k|=r$ and $|y|=s$.
By construction, every word $t\in{\rm Sub}(x)$ with  $|t|\in\{r,s\}$ belongs to $\delta_k^*(x)\subseteq X$
(indeed, we have $r=|x|-qk$ and $s=|x|-rk$).
This implies  $x_ 1,\cdots, x_k,y\in X$, thus $x\in X^{k+1}\cap X$: a contradiction with $X$ being a code.
\cqfd
%
\begin{example} {\rm
\label{EEXX}
(1) According to Proposition \ref{complete-delta-closed}, no code can be $\delta_1$-closed. This can be also drawn from the fact that, for every set $X\subseteq A^+$ we have $\varepsilon\in\delta_1^*(X)$. 

In addition, a code $X\subseteq A^*$ is $\delta_2$-closed if, and only if, it is a subset of $A$.

(2) Let $A=\{a,b\}$ and $k=3$. According to Proposition \ref{complete-delta-closed}, every word in any $\delta_k$-closed code has length not greater than $5$. 
Let $X=\{a^2, ab, b^2,  a^4b, ab^4\}$. We prove that $X$ is a non-complete code which is however maximal as a $\delta_3$-closed code.

Firstly, for proving that $X$ is a code, we apply Sardinas and Patterson  algorithm.
We obtain:  $U_0=X^{-1}X\setminus\{\varepsilon\}=\{a^2b,b^3\}$, 
$U_1=X^{-1}U_0\cup U_0^{-1}X=\{b\}$, 
$U_2=X^{-1}U_1\cup U_1^{-1}X=\{b\}$, whence $U_n=\{b\}$ for every $n\ge 2$, thus $X$  is a code.
Since $\delta_3(X)=\{a^2,ab,b^2\}\subseteq X$, the code $X$ is $\delta_3$-closed.

Secondly, taking for  $\mu$ the uniform Bernoulli distribution, we obtain:
$\mu(X)=3/4+2/32<1$ hence, by Theorem \ref{classic} $X$ is non-complete.

Thirdly, we proceed to  verify that $X$ is maximal in the family of $\delta_3$-closed codes.
For that purpose, by contradiction we assume that a $\delta_3$-closed code $Y$ that strictly contains $X$ exists. 
According to Proposition \ref{complete-delta-closed}, and since $a^4b$ belongs to $Y$, we have $\max\{|y|:y\in Y\}=5$. 
 From the fact that $a^2\in X\subseteq Y$ we have $a\notin Y$ moreover, since $a^4b=(a^2)(a^2)b$, $Y$ cannot contains $b$. Consequently, we have $A\cap Y=\emptyset$ whence,
since $Y$ is  $\delta_3$-closed, no word of length $4$ can belong to $Y$. Similarly, it follows from $\varepsilon\notin Y$ that $Y\cap A^3=\emptyset$: this implies $Y\setminus X\subseteq A^2\cup A^5$.

Note that $\{a^2,ab,ba,b^2\}=A^2$ is a maximal code, therefore $X\cup \{ba\}$, which strictly contains  $A^2$, is not a code, thus we have $ba\notin Y$:
we obtain $Y\setminus X\subseteq A^5$. It follows from $\delta_3(A^5)=A^2$ that no word of $Y\cap A^5$ can contain $ba$ as a subword.
In addition, since we have $a^2,b^2\in X\subseteq Y$, necessarily we have $a^5,b^5\notin Y$, thus  $Y\setminus X\subseteq a^+b^+$. More precisely:

-- Assume $a^3b^2\in Y$. Applying Sardinas and Patterson algorithm to $Y$
leads to compute the sets $U_0$, $U_1$, such that 
$\{a^2b,ab^2,b^3\}\subseteq U_0$ and $\{b^2,b\}\subseteq U_1$. It follows from $b^2\in U_1\cap Y$ that  $Y$ could not be  a code.

-- Similarly by assuming $a^2b^3\in Y$,  applying Sardinas and Patterson algorithm to $Y$
leads to compute the sets $U_0$, $U_1$, which respectively contain the sets $\{a^2b,b^3\}$ and $\{b^2,b\}$. Once more since we have $b^2\in U_1\cap Y$,
$Y$ could not be a code.
As a consequence, no word of $a^+b^+$ can belong to $Y\setminus X$.

Finally we obtain $Y=X$, which is a contradiction: consequently $X$ is maximal in the family of $\delta_3$-closed code over $A$.

} \end{example}
\begin{remark} {\rm
A noticeable fact is that Proposition \ref{complete-delta-closed} provides some bound which is independent of the size of the alphabet, but only depending of $k$.}
 \end{remark}
According to Example \ref{EEXX} (2), there are maximal closed codes that are not complete. In other words no result similar to Theorem \ref{classic4} can be stated in the framework of $\delta_k$-closed codes.
Nevertheless,  the following result holds:
\begin{coro}
\label{coro-delta}
Let $A$ be a  a finite alphabet and let $k\ge1$.
Then one can  decide whether a given non-complete (resp. non-maximal) $\delta_k$-closed code $X\subseteq A^*$ is included into some complete one.
In addition there are a finite number of such complete codes, all of them being computable, if any.
\end{coro}
%
{\flushleft {\it Proof}}
According to Proposition \ref{complete-delta-closed} only a finite number of $\delta_k$-closed codes over $A$ can exist,
each of them being a subset of $A^{\le k^2-k-1}\setminus A^k$.
\cqfd
%

\ \\
In other words, in the framework of $\delta_k$-closed codes we obtain a specific answer  with regard to the open question raised by Lemma \ref{embedding-tau-maximal}.
We close the section by considering  the relation $\iota_k$ and the ones it involves, that is  $I_k$, $S_k$ and $\Lambda_k$:
\begin{proposit}
\label{no-closed-code}
For every every $k\ge 1$, no code can be closed under  $\iota_k$,  nor $I_k$,  $\Delta_k$,  $S_k$,  $\Lambda_k$.
\end{proposit}
%
{\flushleft {\it Proof}}
Let $X\subseteq A^*$ be a $\iota_k$-closed set. According to Lemma \ref{iteration-tau-sur-X}, $X$ is $\iota_k^*$-closed, whence for every 
$x\in X$, the word $x^{k+1}=xx^k\in\iota_k(x)$ belongs to $X$, therefore $X$ cannot be a code.
As a consequence, by definition  no $I_k$-closed code can exist.
According to Example \ref{EEXX}(1), given a code $X\subseteq A^*$, we have $\delta_1(X)\not\subseteq X$: 
this implies $\Delta_k(X)\not\subseteq X$, thus $X$  being not $\Delta_k$-closed, nor  $S_k$-closed, nor $\Lambda_k$-closed.
\cqfd
%
\section{Codes closed under substitutions}
\label{Sec5}
Recall that according to Lemma \ref{iteration-tau-sur-X}, for an arbitrary set, being $\sigma_k$-closed  is  equivalent to being $\sigma_k^*$-closed. Beforehand,  given a word $w\in A^+$, we need a  thorough description of  the set $\sigma_k^*(w)$ (whose any element  of course have length $|w|$). Actually, as shown below, such a set is closely related to the so-called Gray sequences.

\paragraph{Some words about Gray sequences}
\ \\
Binary Gray sequences consist of any  $2^n$-term  sequences of pairwise different words in $A^n$, say $\left(w_i\right)_{1\le i\le 2^n}$, where $A$ is a binary alphabet and $n$ a positive integer, satisfying the following condition:
for each $i\in [1,2^{n}-1]$, the words $w_{i+1}$ and $w_i$ differ by only one letter. Clearly, in the framework of our study, this last condition is equivalent to  $w_{i+1}\in\sigma_1\left(w_i\right)$.
It is well known that, for every positive integer $n$ such sequences exist and they can be computed  by applying now-classical algorithms: see e.g. \citep{G58,JWW80} and for a survey  \citep{S97} or \citep[Chap. 7, Sect. 7.2.1.1]{K05}.
In any case, over a binary alphabet $A$, for every non-empty word $w$, we have $\sigma_1^*(w)=A^{|w|}$.
Furthermore, for every finite alphabet $A$, the so-called $|A|$-arity Gray cyclic sequences themselves allow to generate  $A^n$ \citep{JWW80,R86}: once more we have $\sigma_1^*(w)=A^n$.
In addition, in the special case where  $k=2$ and $|A|=2$, by making use of some Gray sequence, it can be proved that we have $|\sigma_2(w)|=2^{n-1}$  \citep[Exercise 8, p. 28]{K05}.

However,  except for the special cases we mentioned above, to the best of our knowledge, given an arbitrary positive integer $k$ no general description of the structure of $\sigma_k^*(w)$ appears in the literature. 
In any event, in what follows  we provide an exhaustive description of $\sigma_k^*(w)$.
Actually we will see that, according to the fact that $A$ can be a binary alphabet or not, the behavior of $\sigma_k$ greatly differs.

To be more precise, in the case where there are at least three letters in $A$,  the study is greatly  facilitated by the fact that the inclusion  $\sigma_1\subseteq\sigma_k^2$ holds (cf. Lemma \ref{A}).
Unfortunately, this property  does not extend to binary alphabets, but nevertheless,  with this condition the inclusion $\sigma_2\subseteq\sigma_k^2$ holds (cf. Lemma \ref{B}). 
In addition,  in the framework of a binary alphabet, a noticeable fact is that the action of $\sigma_k$ can be translated in terms of some addition on $({\mathbb Z}/2{\mathbb Z})^n$ (cf. Property (\ref{P})).
Let us start by the easiest part of the study.
\subsection{Basic results concerning $\sigma_k^*(w)$: the case where $|A|\ge 3$}
\label{Basic}
In the sequel we set $n=|w|\ge k$. Recall that we set $A^{\ge k}=\bigcup_{i\ge k}A^i$.
We begin with the following property:
\begin{lem}
\label{A}
Assume  $|A|\ge 3$. For every word $w\in A^{\ge  k}$ 
we have $\sigma_1(w)\subseteq \sigma_k^2(w)$.
\end{lem}
%
{\flushleft {\it Proof}}
Recall that  the notation $w=w_1\cdots w_n$,  with $w_i\in A$ ($0\le i\le n$), stands for a  factorization of $w$ upon $A$.
Let $w'\in\sigma_1(w)$; set $w'=w'_1\cdots w'_n$, with $w'_i\in A$ ($0\le i\le n$).
Then a unique $i_0\in [1,n]$, with $n=|w|$,  exists such that:

(a)  $w'_{i}=w_{i}$ if, and only if, $i\neq i_0$.\\
We prove that $w''\in A^*$ exists with
$w''\in \sigma_k(w)$ and $w'\in\sigma_k(w'')$.
It comes from $k\le n$ that some  $(k-1)$-element subset $I\subseteq [1,n]\setminus \{i_0\}$ exists.
Since we have  $|A|\ge 3$, some letter $c\in A\setminus\{w_{i_0},w'_{i_0}\}$ exists.
Let  $w''\in A^n$ such that:

(b) $w''_{i_0}=c$ and, for each $i\neq i_0$: $w''_i\neq w_i$ if, and only if, $ i\in I$.\\
By construction we have $w''\in\sigma_k(w)$, moreover  it comes from $c\neq w'_{i_0}$ that we have $w'_{i_0}\neq w''_{i_0}$.
According to (a) and (b),
we obtain:

(c) $c=w''_{i_0}\neq w'_{i_0}$,

(d) $w'_i=w_i\neq w''_i$ if $i\in I$,  and:

(e)  $w'_i=w_i=w''_i$ if $i\notin I\cup\{i_0\}$.\\
Since we have $|I\cup\{i_0\}|=k$, this implies $w'\in\sigma_k(w'')$. 
\cqfd
%

\ \\
As a consequence of Lemma \ref{A}, in the case where we have $|A|\ge 3$,  the following statement brings some characterization of $\sigma^*(w)$:
\begin{proposit}
\label{A>2-card-Sk*}
Assume $|A|\ge 3$. For each  $w\in A^{\ge k}$, we have $\sigma_k^*(w)=A^{|w|}$.
\end{proposit}
%
{\flushleft {\it Proof}}
Let $w'\in A^n\setminus\{w\}$, with $|w|=n$: we prove that $w'\in\sigma_k^*(w)$.
 Let
$I=\{i_0,\cdots,i_p\}=\{i\in [1,n]: w'_{i}\neq w_i\}$ and
let $\left(w^{(i_j)}\right)_{0\le j\le p}$ be a sequence of words such that both the following conditions hold:

{\rm (a)}  $w=w^{(i_0)}$,~~ $w^{(i_p)}=w'$,

{\rm (b)} for  each $j\in [0,p-1]$, $w^{({i_{j+1}})}_\ell\neq w^{({i_{j}})}_\ell$  if, and only if, $\ell=i_{j+1}$.\\
By construction, the following property holds:

{\rm (c)} for  each $j\in [0,p-1]$,  $w^{({i_{j+1}})}\in\sigma_1\left(w^{({i_{j}})}\right)$ ($1\le j<p$).\\
By induction over $j$ we obtain
$w'\in\sigma_1^*(w)$ thus, according to Lemma \ref{A}: $w'\in\sigma_k^*(w)$. 
\cqfd
%
\subsection{The case of a binary alphabet}
\label{binary}
In the case where $A$ is a binary alphabet, without loss of generality we set $A=\{0,1\}$: this will allow a well-known algebraic interpretation of $\sigma_k$.
Indeed, denote by $\oplus$ the addition in the group ${\mathbb Z}/2{\mathbb Z}$ with identity $0$,
and  fix a positive integer $n$. Let $w=w_1\cdots w_n$, $w'=w'_1\cdots w'_n$, with $w_i,w'_i\in A$ ($1\le i\le n$).
Define $w\oplus w'$ as the unique word of $A^n$ such that, for each $i\in [1,n]$, the letter of position $i$ in $w\oplus w'$ is $w_i\oplus w'_i$. 
With this notation the sets $A^n$ and $({\mathbb Z}/2{\mathbb Z)}^n$ are in one-to-one correspondence.

From the previous remarks, we have $w'\in\sigma_1(w)$ if, and only if, some $u\in A^n$ exists such that  $w'=w\oplus u$ with  $|u|_1$,
the number of occurrences of the letter $1$ in $u$, equal to $1$
 (equivalently, we have  $|u|_0=n-1$).
From the fact that we have $\sigma_k(w)\subseteq\sigma_1^k(w)$,  the following property holds:
\begin{eqnarray}
\label{P}
w'\in\sigma_k(w)\Longleftrightarrow \exists u\in A^n: w'=w\oplus u,~~|u|_1=k.  
 \end{eqnarray}
More precisely, for each $i\in [1,n]$, the condition $u_i=1$ is equivalent to $w_i\neq w'_i$.
Let $d=|\{i\in [1,n]: w_i=w'_i=1\}|$.  On a first hand, it follows from  $|u|_1=|\{i\in [1,n]:w_i=1,w'_i=0\}|+|\{i\in [1,n]:w_i=0,w'_i=1\}|$, that 
$|u|_1=(|w|_1-d)+(|w'|_1-d)=|w|_1+|w'|_1-2d$, thus  $|u|_1=|w|_1+|w'|_1\bmod 2$. On another hand, we have  $|w'|_1-|w|_1=|w'_1|+|w|_1-2|w|_1$,
thus $|w'|_1-|w|_1=|w_1|+|w'|_1\bmod 2$.
We obtain:
\begin{eqnarray}
\label{Q}
 w'=w\oplus u \Longrightarrow |w|_1+|w'|_1=|w_1|-|w'|_1\bmod{2}=|u|_1\bmod{2}.
 \end{eqnarray}
In addition $w'=w\oplus u$ is equivalent to $u=w\oplus w'$.
Finally, for $a\in A$ we denote by ${\overline   a}$ its complementary letter that is, we set ${\overline   a}=a\oplus 1$;
moreover, for $w=w_1\cdots w_n$, with $w_i\in A$ ($i\in [1,n]$), we set ${\overline   w}={\overline   w_1}\cdots {\overline   w_n}$.
The following statement is the counterpart of  Lemma \ref{A} in the framework of binary alphabets:
\begin{lem}
\label{B}
Assume $|A|=2$. For every $w\in A^{\ge k+1}$, 
we have $\sigma_2(w)\subseteq \sigma_k^2(w)$.
\end{lem}
%
{\flushleft {\it Proof}}
Set $A=\{0,1\}$. It follows from $\sigma_2\subseteq\sigma_1^2$ that the result holds for $k=1$: in the sequel of the proof, we assume $k\ge 2$.  Let $n=|w|\ge k+1$ and $w'\in \sigma_1(w)$.
Set   $w=w_1\cdots w_n$,  $w'=w'_1\cdots w'_n$, with $w_i,w'_i\in A$ ($1\le i\le n$).
Note that we have  $w_i\neq w'_i$ if, and only if, the equation  $w_i=\overline   w'_i$ holds.
By construction, there are distinct integers $i_0,j_0\in [1,n]$  such that the following condition holds for each $i\in [1,n]$:

(a) $w'_i=\overline  {w_i}$ if, and only if, $i\in\{i_0,j_0\}$.\\
%
It follows from $n\ge k+1\ge 3$ that some $(k-1)$-element set $I\subseteq [1,n]\setminus \{i_0,j_0\}$ exists. 
Let $w'', w'''\in A^n$ such that each of the two following conditions holds: 

(b) $w''_{i}=\overline {w_{i}}$ if, and only if, $i\in \{i_0\}\cup I$, and:

(c) $w'''_{i}=\overline {w''_i}$ if, and only if, $i\in\{j_0\}\cup I$.\\
%
By construction, we have  $w'''\in\sigma_k(w'')$ and $w''\in\sigma_k(w)$, thus $w'''\in\sigma_k^2(w)$.
Moreover, the fact that we have $w'''=w'$ is attested by the three following equations:

(d) $w'''_{j_0}=\overline {w''_{j_0}}=\overline {w_{j_0}}=w'_{j_0}$,

(e) $w'''_{i_0}=w''_{i_0}=\overline {w_{i_0}}=w'_{i_0}$, and:

(f) for $i\notin \{i_0,j_0\}$: $w'''_{i}=\overline {w''_{i}}=w_i=w'_i$ if, and only if, $i\in I$.
\cqfd
%

\ \\
As regards   algebraic interpretation of binary alphabets, we state: 
\begin{lem}
\label{C}
Let $A=\{0,1\}$. Given $w,w'\in A^n$ each of the two following properties holds:

{\rm (i)} If we have $|w|\ge k+1$ and  $w'\in\sigma_k^*(w)$,  where $k$ is  even, then $|w'|_1-|w|_1$ is an even integer;

{\rm (ii)} If $|w'|_1-|w|_1$ is even then we have $w'\in\sigma_k^*(w)$, for every $k$ such that $|w|\ge k+1$.
\end{lem}
%
{\flushleft {\it Proof}}
Assume $k$ even  with $w'\in\sigma_k^*(w)$.
According to Property (\ref{P})  we have $w'=w\oplus u$ with $|u|_1=k$.
According to Property (\ref{Q}), $|w'|_1-|w|_1$ is  even, hence Property (i) holds.

Conversely, assume $|w'|_1-|w|_1$ even
and let  $u=w\oplus w'$. According to Property (\ref{Q}),   $|u|_1$ is an even integer: set  $|u|_1=2p$, with $p\ge 0$. Actually we have  $u=u^{(1)}\oplus\cdots\oplus u^{(p)}$, with $|u^{(i)}|_1=2$ for each $i\in [1,p]$,
and the sets $D_i=\{j: u^{(i)}_j=1\}$ ($1\le i\le p$) being pairwise disjoint.
Let $\left(w^{(0)},\cdots, w^{(p)}\right)$ be the sequence of words in $A^n$ defined by $w^{(0)}=w$, $w^{(p)}=w'$ and $w^{(i)}=w^{(i-1)}\oplus u^{(i)}$ ($1\le i\le p$).
 For each $i\in [1,p]$, by taking $k=2$ in Property (2) we obtain  $w^{(i)}\in\sigma_2(w^{(i-1)})$.
By induction, since  the sets $D_{i'}$  ($1\le i'\le p$) are pairwise disjoint, this implies $w^{(i)}\in\sigma_2^i(w^{(0)})$: in particular we have $w'\in\sigma_2^p(w)$.
According to Lemma \ref{B}, we obtain $w'\in\sigma_k^*(w)$  for every $k\le |w|-1$: this establishes Property (ii).
\cqfd
%

\ \\
Given a positive integer $n$, we denote ${\rm Even}_1^n$ (resp., ${\rm Odd}_1^n$) the set of the words $w\in A^n$ such that $|w|_1$ is even (resp., odd). 
As a consequence of Lemma \ref{B} and Lemma \ref{C}, we state:
\begin{proposit}
\label{card-Sk*}
Assume $|A|=2$. For each word $w\in A^{\ge k}$ exactly one of  the following conditions holds:

{\rm (i)}  $|w|\ge k+1$,  $k$ is even, and $\sigma_k^*(w)\in\{{\rm Even}_1^{|w|},{\rm Odd}_1^{|w|}\}$;

{\rm (ii)} $|w|\ge k+1$, $k$ is odd, and  $\sigma_k^*(w)=A^{|w|}$;

{\rm (iii)} $|w|=k$ and $\sigma_k^*(w)=\{w,{\overline  w}\}$.
\end{proposit}
%
{\flushleft {\it Proof}}
Let $w\in A^{\ge k}$ and $n=|w|$. Trivially, the case  where $n=k$ corresponds to Condition (iii) of the statement.

Next, we assume $n\ge k+1$, with $k$ even. It follows from  Lemma \ref{C}(i) that  $\sigma_k^*(w)$ is the set of the words  $w'\in A^n$ such that $|w'|_1- |w|_1$ is even: this corresponds to Condition (i).

At last, we assume $n\ge k+1$ and  $k$ odd. We will prove that we have $w'\in\sigma_k^*(w)$ for each word $w'\in A^n\setminus\{w\}$.  If $|w'|_1-|w|_1$ is  even, the property comes from  Lemma \ref{C}(ii). Assume $|w'|_1-|w|_1$ odd and let $t\in\sigma_1(w')$ that is, $w'\in\sigma_1(t)\subseteq\sigma_{k}\left(\sigma_{k-1}(t)\right)$ thus, $w'\in\sigma_k(t')$ for some $t'\in\sigma_{k-1}(t)$. According to Property (\ref{P}), it follows from $w'\in\sigma_1(t)$ that $|t|_1-|w|'_1$ is odd, whence $|t|_1-|w|_1=(|t|_1-|w'|_1)+(|w'|_1-|w|_1)$ is even: according to Lemma \ref{C}(ii), this implies $t\in\sigma^*_{k}(w)$. But since $k-1$ is even, we have $\sigma_{k-1}(t)\subseteq\sigma_2^*(t)$, thus $t'\in\sigma_2^*(t)$: according to  Lemma \ref{B}, this implies $t'\in\sigma_{k}^*(t)$ (we have $|t|=|w'|=n\ge k+1$). We obtain $w'\in\sigma_k(t')\subseteq\sigma_{k}^*(t)\subseteq \sigma_k^*\left(\sigma_k^*(w)\right)=\sigma_k^*(w)$: this completes the proof of  Condition (ii).
\cqfd
%
\subsection{The consequences for $\sigma$-closed codes}
\label{Cons}
Let $X\subseteq A^*$ be a $\sigma_k$-closed code. Beforehand, we notice that it may happen that the inclusion $X\subseteq A^{\le k-1}$ holds: indeed, trivially every subset of $A^{\le k-1}$ is $\sigma_k$-closed. In the case where at least one word  in $X$, say $x$,  has length not smaller than $k$, thanks to the study we have drawn in both the sections \ref{Basic} and  \ref{binary}, we are  able to describe $\sigma_k^*(x)$.
The aim of Section \ref{Cons} is to apply such a study in order to precisely describe the structure of our code $X$.

More precisely, in the two special cases where we have $|A|\ge 3$, or $|A|=2$ with $k$ odd, due to the fact that the equation $\sigma_k^*(x)=A^{|x|}$ holds, we will see that the structure of $X$ can be described in a straightforward way (cf. Lemma \ref{Code-sigma-fini}, set out below). Actually, the most delicate part of the study consists in  examining the case where we have  $|A|=2$ and $k$ even: this corresponds to Condition (\ref{D}), which is stated just below. With such a condition, by making use of  some technical property (cf. Lemma  \ref{v-sk*w}), an exhaustive description of the structure of the code $X$ can be obtained  (cf. Lemma \ref{sigma**-code}).
At last, some summary of the study is provided by Corollary \ref{Sigma-k-closed}. Let us start by stating the announced condition:

\bigskip
Given a $\sigma_k$-closed code $X\subseteq A^*$, we say that the tuple $(k,A,X)$ satisfies Condition (\ref{D}) if each of the three following properties holds:
\begin{eqnarray}
\label{D} 
{\rm (a)}~k~{\it is~even,}~~~{\rm (b)}~ |A|=2, ~~~{\rm (c)}~X\not\subseteq A^{\le k}.
\end{eqnarray}
At first, we establish the following property:
\begin{lem}
\label{v-sk*w}
Assume  $|A|=2$ and $k$ even.  Given a pair of words  $v,w\in A^+$, 
if  we have $|w|\ge \max\{|v|+1,k+1\}$ then  the set $\sigma_k^*(w)\cup \{v\}$ cannot be  a code.
\end{lem}
%
{\flushleft {\it Proof}}
Let $v,w\in A^+$ and $n=|w|\ge \max\{|v|+1,k+1\}$: we have $v\notin\sigma_k^*(w)\subseteq A^{n}$.
We are in Condition (i) of Proposition \ref{card-Sk*} that is, we have  $\sigma_k^*(w)\in\{{\rm Even}_1^n,{\rm Odd}_1^n\}$.

On a first hand, since  $A^{n-1}$ is a right-complete prefix code \citep[Theorem 3.3.8]{BPR10}, it follows from $|v|\le n-1$ that a (perhaps empty) word $s$ exists such that $vs\in A^{n-1}$. 
On another hand, it  follows from $A^{n-1}A=A^n={\rm Even}_1^n\cup {\rm Odd}_1^n$ that, for each $u\in A^{n-1}$, a unique pair of letters $a_0,a_1$, exists such that $ua_0\in {\rm Even}_1^n$, $ua_1\in {\rm Odd}_1^n$ with $a_1=\overline  {a_0}$.

In other words,
$a\in A$ exists such that $vsa\in \sigma_k^*(w)$.
According to Lemma \ref{C}(i), the integer $|sav|_1-|w|_1=|vsa|_1-|w|_1$ is even; according to Lemma \ref{C}(ii), this implies $sav\in\sigma_k^*(w)$. Since we have   $(vsa)v=v(sav)$, the set $\sigma_k^*(w)\cup \{v\}$ cannot be a code.
\cqfd
%
\ \\

As a consequence of Lemma \ref{v-sk*w}, we obtain the following result:
\begin{lem}
\label{sigma**-code}
Given a 
$\sigma_k$-closed code $X\subseteq A^*$,
if $(k,A,X)$ satisfies Condition (\ref{D}) 
then
we have  $X\in\{{\rm Even}_1^n,{\rm Odd}_1^n,A^n\}$, for some $n\ge k+1$.
\end{lem}
%
{\flushleft {\it Proof}}
 Firstly, by contradiction assume that two words $x,y\in X\cap A^{\ge k+1}$ exist such that
 $|x|\neq |y|$ that is,  
without loss of generality $|x|\ge |y|+1$.
Since $X$ is $\sigma_k$-closed, we have $\sigma_k^*(x)\subseteq X$.
Since every subset of a code is a code, the subset of $X$, $\sigma_k^*(x)\cup\{y\}$, is a
code as well, thus contradicting the result of Lemma
\ref{v-sk*w}.
As a consequence, all the words in $X\cap A^{\ge k+1}$ have a common length that is, we have $X\subseteq A^{\le k}\cup A^n$, for some integer $n\ge k+1$.

Secondly, once more by contradiction, assume that there are  words  $x\in X\cap A^{\ge k+1}$, $y\in  X\cap A^{\le k}$.
As indicated above, since $X$ is $\sigma_k$-closed, $\sigma_k^*(x)\cup\{y\}$, which is a subset of $X$, is a code:
since we have $|x|\ge k+1$ and $|x|\ge |y|+1$, once more we obtain a contradiction with the result of Lemma \ref{v-sk*w}.

As a consequence,  either we have $X\subseteq A^{\le k}$ or we have $X\subseteq A^{\ge k+1}$, for some $n\ge k+1$: since $(k,A,X)$ satisfies Condition (\ref{D}), only the second condition holds.
According to Proposition \ref{card-Sk*}(i),  for each word $x\in X$ we obtain $\sigma_k^*(x)\in \{{\rm Even}_1^n,{\rm Odd}_1^n\}$. 
It follows from $\sigma^*(X)\subseteq X$ that we have either ${\rm Even}_1^n\subseteq X$, or ${\rm Odd}_1^n\subseteq X$, or $A^n\subseteq X$. 
We now examine each of these three conditions:

--  Since $A^n$ is a maximal code, the condition $A^n\subseteq X$ implies $X=A^n$.

-- Now, we examine the case where the condition  ${\rm Even}_1^n\subseteq X$ holds. 
Assume  that  we have ${\rm Even}_1^n\neq X$ that is, some word $x\in X\cap {\rm Odd}_1^n$ exists.
Since $k$ is an even integer, once more according to Proposition \ref{card-Sk*}(i), we have $\sigma_k^*(x)={\rm Odd}_1^n$. 
On a first hand we have $\sigma_k^*({\rm Even}_1^n\cup\{x\})= {\rm Even}_1^n\cup \sigma_k^*(x)={\rm Even}_1^n\cup {\rm Odd}_1^n=A^n$, furthermore ${\rm Even}_1^n\subseteq X$ implies $X\in\{{\rm Even}_1^n,A^n\}$.
On another hand we have $\sigma_k^*({\rm Even}_1^n\cup\{x\})\subseteq\sigma_k^*(X)\subseteq X$: 
since $A^n$ is a maximal code, once more we obtain $X=A^n$. Consequently,  in any case, ${\rm Even}_1^n\subseteq X$ implies  $X\in\{{\rm Even}_1^n,A^n\}$

-- Symmetrical arguments prove that the condition  ${\rm Odd}_1^n\subseteq X$ implies $X\in\{{\rm Odd}_1^n,A^n\}$.\\ 
 Consequently in any case we have 
$X\in\{{\rm Even}_1^n,{\rm Odd}_1^n,A^n\}$: this completes the proof.
\cqfd
%

\ \\
According to Lemma \ref{sigma**-code}, with Condition \ref{D}, no $\sigma_k$-closed code can simultaneously possess words in $A^{\le k}$ and words in $A^{\ge k+1}$.
In remains  to examine the case where  Condition (\ref{D}) does not hold. The following property allows to complete this part of the study:
\begin{lem}
\label{Code-sigma-fini}
Given a  $\sigma_k$-closed code $X\subseteq A^*$, if $(k,A,X)$ does not satisfy Condition (\ref{D}) then
either we have $X\subseteq A^{\le k}$, or  we have $X=A^n$, with  $n\ge k+1$.
\end{lem}
%
{\flushleft {\it Proof}}
Assume that Condition (\ref{D}) doesn't hold.  By definition, exactly  one of the three following conditions holds:

(a)  $X\subseteq A^{\le k}$;

(b) $X\not\subseteq A^k$ and $|A|\ge 3$;

(c)  $X\not\subseteq A^{\le k}$ with $|A|=2$ and $k$ odd.

\smallskip
With each of the two last conditions, let $x\in X\cap A^{\ge k+1}$. 
Since $X$ is $\sigma_k$-closed, according to the propositions \ref{A>2-card-Sk*} and \ref{card-Sk*}(ii), we have $A^n=\sigma_k^*(x)\subseteq\sigma_k^*(X)\subseteq X$. 
Since $A^n$ is a maximal (bifix) code, we obtain
$X=A^n$. 
\cqfd

\ \\
%
At last, as a consequence of Lemma \ref{sigma**-code}  and Lemma \ref{Code-sigma-fini}, we state:
\begin{coro}
\label{Sigma-k-closed} Let $A$ be a finite alphabet and $k$ a positive integer. Given a code $\sigma_k$-closed $X\subseteq A^*$, either $X$ is a subset of in $A^{\le k}$, or we have $X\in\{{\rm Even}_1^n,{\rm Odd}_1^n,A^n\}$ for some integer $n\ge k+1$.
\end{coro}
%
{\flushleft {\it Proof}}
Let $x\in X$ and $n=|x|$.
As specified in preamble of Section \ref{Sec5}, we have $\sigma_1^*(x)=A^{n}$  (see e.g. \citep[Chap. 7, Sect. 7.2.1.1]{K05}).
Consequently, the property  of Corollary \ref{Sigma-k-closed} holds in the case where $k=1$. In the sequel of the proof, we assume $k\ge 2$.
Assume that  $X$ is $\sigma_k$-closed.
According to Lemma \ref{Code-sigma-fini}, if Condition \ref{D} does not hold, the code $X$ satisfies our property. 
Otherwise, according to  Lemma \ref{sigma**-code} we have $X\in\{{\rm Even}_1^n,{\rm Odd}_1^n,A^n\}$, whence the code $X$ once more satisfies the property.
\cqfd

\subsection{Maximality and completeness in  $\sigma_k$-closed codes}
We are now ready to provide an exhaustive description of complete $\sigma_k$-closed (resp., $\Sigma_k$-closed) codes:
\begin{proposit}
\label{sigma*-complete}
Let $X\subseteq A^*$ a code. Then each of the following properties holds:

{\rm (i)} If $X$ is $\sigma_k$-closed and complete, then either
$X$ is a subset of $A^{\le k}$, or  some integer $n\ge k+1$ exists such that $X=A^n$.

{\rm (ii)} If $X$ is $\Sigma_k$-closed, we have $X=A^n$ for some $n\ge k$, thus it is necessarily maximal and complete.
\end{proposit}
%
{\flushleft {\it Proof}}
Let $X$ be a   complete $\sigma_k$-closed code. 
According to Corollary \ref{Sigma-k-closed}, either we have $X\subseteq A^{\le k}$, or we have  $X\in\{{\rm Even}_1^n,{\rm Odd}_1^n,A^n\}$ for some integer $n\ge k+1$. Taking for $\mu$ the uniform Bernoulli distribution, we have $\mu({\rm Even}_1^n)=\mu({\rm Odd}_1^n)=1/2$, and $\mu(A^n)=1$, thus according to Theorem \ref{classic},$X=A^n$.

In view of  Property (ii), recall that by definition we have   $\Sigma_k(X)=\bigcup_{1\le i\le k}\sigma^i_k(X)$: this implies  $\sigma_1(X)\subseteq \Sigma_k(X)$.
Consequently, given a $\Sigma_k$-closed code $X$, some integer $n\ge 1$ exists such that  $A^n= \sigma_1^*(X)\subseteq \Sigma_k^*(X)\subseteq X$. Since  $A^n$ is a maximal code we obtain  $X=A^n$, whence $X$ is complete. \cqfd
%

\ \\
Trivially, according to Proposition \ref{sigma*-complete}(ii), in the family of $\Sigma_k$-closed codes maximality and completeness are equivalent notions.
In addition, as a  direct consequence of  Proposition \ref{sigma*-complete} (i), 
in the family of $\sigma_k$-closed codes  included in $A^{\ge k+1}$, those concepts are also equivalent. 

With regard to $\sigma_k$-closed codes not included in $A^{\ge k+1}$, results  are different. On a first hand,  according to Proposition \ref{sigma*-complete}(i), such codes are necessarily included in $A^{\le k}$.
On another hand, as shown in \citep{R77}, there are non-complete finite codes that cannot be included into any finite complete (or equivalently, finite maximal) one. 
Let $X$ be one of them  and let $k=\max\{|x|:x\in X\}$+1.
By definition $X$ is $\sigma_k$-closed. Since every $\sigma_k$-closed code is finite, no finite maximal code can contain $X$; in other words, although $X$ is  non-complete, it  is  maximal in the family of $\sigma_k$-closed codes. 
\begin{example} {\rm\citep{R77}
Let $A=\{a,b\}$ and $X=\{a^5,a^2ba,a^2b,ba,b\}$, $k=6$. The code $X$ is non-complete, $\sigma_k$-closed and not included into any finite maximal code,
whence $X$ is maximal in the family of $\sigma_k$-closed codes.
} \end{example}
\begin{proposit}
\label{sigma-completion}
Let $X$ be a (finite) non-complete $\sigma_k$-closed code. Then  one can decide whether some complete $\sigma_k$-closed code containing $X$ exists.
More precisely,  there is only a finite number of such codes, each of them being computable, if any.
\end{proposit}
%
{\flushleft {\it Proof}}
We draw  the scheme of an algorithm that allows to compute every complete $\sigma_k$-closed code ${\hat X}$ containing $X$.

-- In a first step, we compute $Y=X\cap A^{\le k}$.
 
-- If $Y=X$,  according to Proposition \ref{sigma*-complete}, we have  ${\hat X}\subseteq A^{\le k}$: ${\hat X}$, if any,  can be computed in a finite number of steps.

-- Otherwise, ${\hat X}$ exists if, and only if, for some $n\ge k+1$ we have $X\subseteq A^n$: this can be checked in a straightforward way; furthermore we obtain $\hat X=A^n$.
\cqfd
%

\ \\
Recall that  Corollary \ref{coro-delta} has provided some method for embedding a $\delta_k$-closed code into some maximal one (if any). Similarly, in  the framework of $\sigma_k$-closed codes, Proposition \ref{sigma-completion} actually brings a positive answer to the issue raised by Lemma \ref{embedding-tau-maximal}.
\section{Some future line of research}
\label{Last}

The study we presented in  the present paper lies in the framework of the free monoid, and it involves some connections with the three famous fields of error detection, regular binary relations, and variable-length codes. With regard to further developments, such connections appear promising:
%
\begin{enumerate}[label={\rm (\roman*)}]
\item On a first hand, as regards independence of codes,  the constraints  introduced in Section \ref{Sec4} lead to some regard  of the framework of error detection in term of free monoid. From this point of view, investigations could be done in several ways:

-- According to Lemma \ref{embed0}, every  code independent with respect to a given edit relation can be embedded into some maximal one. 
We recall that presently there is no method of computation, as is the case of the formula provided by Theorem \ref{EhrRoz}.
Developing such methods, at least for special families of codes  could allow new  connections between variable-length codes and error-detecting (error-correcting) ones.

-- As attested by the examples of Section \ref{EXEM}, it appears very difficult to construct codes that satisfies the totality of the constraints \ref{0}--\ref{4}  of Section \ref{EDC}.
Fortunately, alternative solution exist in order to satisfying the condition of error correction.
From this point of view, according to the type of channel that is, the type of edit relation, it would be desirable to identify noticeable families of regular (even finite) variable-length codes that could as to best  ensure error correction constraint.

-- Studying whether the questions  we stated in Section \ref{decidable0} are decidable or not, appears challenging. 
From this last point of view, new connections between regular binary words relations and variable-length codes (especially maximal ones) could be brought to light.

\item
On another hand, with  regard to closed codes, 
according to the results of the propositions \ref{A>2-card-Sk*} and  \ref{card-Sk*} 
one can ask whether some sequences generalizing the classical Gray sequences exist in $A^n$, or eventually in the sets ${\rm Even}_1^n$ or ${\rm Odd_1}^n$.
In such sequences, two consecutive elements would differ by exactly $k$ characters.
Cyclic sequences that is, sequences $(w_i)_{1\le i\le p}$ such that $w_1=\sigma_k(w_p)$, would be highly desirable: indeed, such a property is satisfied by  each of the Gray sequences provided by  the literature.
Actually, in view of  some of our most recent studies, we strongly believe that the answer is yes.  We hope to develop this point in some further paper.
\item
At least, it could be of interest to extend the study of the present paper to the framework of other  specific binary relations that is, other specific  (quasi) metrics.
\end{enumerate}
\section*{Declaration of competing interest:} 
None.
\section*{Acknowledgments}
We are grateful to the anonymous reviewers for thorough examination of the paper, and fruitful suggestions and comments.
 \renewcommand{\bibsection}{\section*{References}}
\bibliography{mysmallbib}
\bibliographystyle{plain}
\makeatother
\end{document}